\documentclass[lettersize,journal]{IEEEtran}
\usepackage{amsmath,amsfonts}
\usepackage[ruled,linesnumbered]{algorithm2e}
\usepackage{array}
\usepackage{textcomp}
\usepackage{stfloats}
\usepackage{url}
\usepackage{verbatim}
\usepackage{graphicx}
\usepackage{cite}
\usepackage{booktabs}
\usepackage{graphbox}
\usepackage{caption}
\usepackage{subcaption}
\usepackage{hyperref}
\usepackage{multirow}
\usepackage{balance}
\usepackage[dvipsnames,table]{xcolor}

\usepackage{pifont}
\newcommand{\cmark}{\textcolor{ForestGreen}{\ding{51}}}%
\newcommand{\xmark}{\textcolor{red}{\ding{55}}}%

\usepackage[acronym]{glossaries}

\makeglossaries
\glsdisablehyper
\newacronym{mdp}{MDP}{Markov decision process}
\newacronym{pomdp}{POMDP}{partially observable Markov decision process}
\newacronym{rl}{RL}{reinforcement learning}
\newacronym{ppo}{PPO}{proximal policy optimization}
\newacronym{drl}{DRL}{deep reinforcement learning}

\newacronym{uav}{UAV}{unmanned aerial vehicle}
\newacronym{nfz}{NFZ}{no-fly zone}
\newacronym{cpp}{CPP}{coverage path planning}
\newacronym{fov}{FoV}{field of view}

\usepackage{tikz}
\usetikzlibrary{arrows.meta}
\usetikzlibrary{positioning}

\begin{document}

\title{Learning to Recharge: UAV Coverage~Path~Planning through Deep Reinforcement Learning}

\author{Mirco Theile,
Harald Bayerlein, 
Marco Caccamo, %
Alberto L. Sangiovanni-Vincentelli
\thanks{Marco Caccamo was supported by an Alexander von Humboldt Professorship endowed by the German Federal Ministry of Education and Research.}%
\thanks{Mirco Theile, Harald Bayerlein, and Marco Caccamo are with the TUM School of Engineering and Design at the Technical University of Munich, Germany. E-mail: \{mirco.theile, h.bayerlein, mcaccamo\}@tum.de}
\thanks{Mirco Theile and Alberto L. Sangiovanni-Vincentelli are with the Department of Electrical Engineering and Computer Science at the University of California Berkeley, USA. E-mail: alberto@berkeley.edu}
\thanks{This work has been submitted to the IEEE for possible publication. Copyright may be transferred without notice, after which this version may no longer be accessible.}}

\maketitle

\begin{abstract}
Coverage path planning (CPP) is a critical problem in robotics, where the goal is to find an efficient path that covers every point in an area of interest. This work addresses the power-constrained CPP problem with recharge for battery-limited unmanned aerial vehicles (UAVs). In this problem, a notable challenge emerges from integrating recharge journeys into the overall coverage strategy, highlighting the intricate task of making strategic, long-term decisions. We propose a novel proximal policy optimization (PPO)-based deep reinforcement learning (DRL) approach with map-based observations, utilizing action masking and discount factor scheduling to optimize coverage trajectories over the entire mission horizon. We further provide the agent with a position history to handle emergent state loops caused by the recharge capability. Our approach outperforms a baseline heuristic, generalizes to different target zones and maps, with limited generalization to unseen maps. We offer valuable insights into DRL algorithm design for long-horizon problems and provide a publicly available software framework for the CPP problem. 
\end{abstract}

\section{Introduction}

Determining a path that covers every point in an area of interest is the designated goal of the \gls{cpp} problem. Typically, this path should be as short and efficient as possible, making the problem related to the traveling salesman problem~\cite{choset2001}. At the same time, other constraints, e.g., collision avoidance, need to be satisfied. When a complete geometric description of the area of interest is available, the CPP problem is proven to be NP-hard~\cite{arkin2000approximation}. Rapid innovation has enabled \glspl{uav}, or drones as they are commonly called, to be used for a wide range of applications that constitute some form of the CPP problem, e.g., smart farming, photogrammetry, disaster management, environmental monitoring, or infrastructure inspection~\cite{cabreira2019}. Including but not limited to these applications, the overall UAV services market is projected to reach a value of USD 189.4 billion by 2030~\cite{droneMarket2023}. When using autonomous quadcopter-type \glspl{uav} for the \gls{cpp} problem, a central constraint is the limited onboard battery capacity that may make repeated recharge cycles necessary to complete a coverage mission.

As a long-standing problem in robotics, numerous path planning algorithms have been proposed to solve various instances of the CPP problem. Most classically, these approaches are either characterized as complete approaches or heuristics~\cite{choset2001,galceran2013}. Complete approaches guarantee complete coverage of all free space in the scenario, but may be computationally infeasible, while heuristics do not offer guarantees but are usually computationally simpler. Algorithms of both categories are often based on cellular decomposition of the environment, i.e., a geometric partition into cells that either need to be covered or are seen as obstacles. Each cell is then covered by some strategy, e.g., simple back-and-forth motions, while the order in which the cells are visited is decided by some planning algorithm. Previous works have suggested the use of graph search~\cite{xu2011optimal}, dynamic programming~\cite{xie2018integrated}, meta-heuristics~\cite{valente2013aerial}, or evolutionary algorithms~\cite{bouzid2017quadrotor} to plan coverage paths for UAVs. 

In this work, we formulate the \textit{power-constrained CPP problem with recharge}. In contrast to most existing problem formulations, we consider an energy-limited case in which a UAV can achieve full coverage of the target region only by carrying out multiple flights with intermediate charging stops. We only consider a coverage mission completed when full coverage of the target region is achieved. Furthermore, we investigate a more generalized form of the CPP problem, where only a subset of the free space is considered a coverage target. This challenging scenario requires optimization over the entire time horizon of the mission, with a complex interplay between flight segments to and from landing zones. 

We propose to address this problem using \glsentryfull{drl}, as it has the potential to find close to optimal solutions for this NP-hard problem. To that end, we formulate the problem such that the \gls{ppo}-based~\cite{schulman2017proximal} \gls{drl} agent observes the problem as global and local maps~\cite{theile2021uav} and learns to minimize the steps required to solve the problem. We utilize \textit{action masking} based on a safety model, which guarantees safety and significantly improves the learning performance of the agent. We address the long-horizon problem by scheduling the discount factor, which balances the importance of immediate and future rewards in the objective. The scheduling allows the agent to first learn to solve the task, followed by continuous improvement on the efficiency of the trajectory. Furthermore, we provide the agent with a history of its positions to aid it in solving an emergent state loop problem that is caused by the recharge capability and the consequential possibility of recreating previously visited states. In the results, we conduct ablation studies on action masking, discount factor scheduling, and position history. Additionally, we highlight the agent's zero-shot generalization ability, denoting its capacity to solve unseen scenarios without additional training. We evaluate this generalization across various target zones and maps, including its capability to solve unseen maps. As existing approaches from the literature can usually not be easily adapted to the power-constrained CPP problem with recharge, we further provide a model-based greedy heuristic and compare it to our proposed algorithm, showcasing that our \gls{drl} approach can reliably find shorter paths to solve the CPP problem.

\gls{cpp} is a fundamental problem with broad applications, holding the potential to address numerous existing challenges across various domains. However, only a few papers investigated DRL-based methods specifically tailored to the complexities of the UAV coverage problem~\cite{theile2021uav, theile2020uav, dong2021collaborative}, or investigated the general applicability of DRL for ground-based CPP~\cite{chen2019adaptive, kyaw2020coverage, jonnarth2023end, heydari2021reinforcement, apuroop2021reinforcement, lakshmanan2020complete, nasirian2021efficient}. In this work, we present several contributions to address these gaps:
\begin{itemize}
    \item We tackle the power-constrained CPP problem with recharge, a critical aspect for battery-limited UAVs operating in large and complex environments, utilizing a PPO-based global-local DRL approach.
    \item Using model-based action masking, we ensure the safety of the generated flight paths.
    \item Through extensive evaluations, we show that our approach outperforms a baseline heuristic, can generalize planning over multiple maps with different characteristics, and can even solve a previously unseen map.
    \item We offer valuable design insights for developing DRL algorithms and corresponding neural network models for long-horizon problems.
    \item Finally, we provide a publicly available software framework\footnote{\url{https://github.com/theilem/uavSim.git}} based on the commonly used OpenAI gym API~\cite{gymlibrary} and include all trained models shown in this work, filling the current gap in available software tools.
\end{itemize}

This paper is organized as follows: Section~\ref{sec:related_work} provides an overview of relevant literature in coverage path planning and DRL approaches. In Section~\ref{sec:problem}, we introduce the full CPP problem with recharge and associated environmental constraints. Starting from the problem description as a partially observable Markov decision process (POMDP), we introduce our DRL-based solution methodology in Section~\ref{sec:methodology}. Section~\ref{sec:simulation}, which describes our evaluation setup, is followed by a detailed analysis of results and ablation studies in Section~\ref{sec:results}. We conclude the paper with a summary and an outlook to future work in Section~\ref{sec:conclusion}.

\section{Related Work}
\label{sec:related_work}
As CPP is a long-standing problem in robotics, a wide array of solution methods has been proposed. The most classical introduction to the CPP problem and traditional solution methods can be found in~\cite{choset2001}. Cabreira et al.~\cite{cabreira2019} provide a survey focusing specifically on decomposition-based CPP approaches for UAVs. In contrast, Tan et al.~\cite{tan2021comprehensive} review existing CPP approaches (not specifically for UAVs) ranging from decomposition-based, exact algorithms to fully random heuristics, including recent deep learning-based approaches. 

Focusing on designing coverage paths for UAVs specifically, most previous works suggest non-DRL approaches. An example of planning a coverage path for a UAV through a combination of a genetic algorithm and the rapidly exploring random tree (RRT) method is given in~\cite{bouzid2017quadrotor}, which leads to an optimal path for the specific scenario but is only practical for small maps and simple environments. In~\cite{xie2018integrated}, the authors suggest solving the underlying traveling salesman problem of flying from one region of interest to the next with dynamic programming and using a straightforward back-and-forth motion to cover the individual regions. They focus on an environment without obstacles and the necessity to recharge. Meta-heuristics like harmony search in~\cite{valente2013aerial}, specifically for a UAV application scenario in precision agriculture, offer relatively low computation time but without optimality guarantees. Various variants of posing a CPP problem in a graph structure by decomposition and solving it by graph search algorithms have also been suggested. In~\cite{xu2011optimal}, such an algorithm is developed and evaluated on a real-world UAV system. None of these approaches would be directly applicable to the scenario we investigate, be it for lack of recharging, lower complexity environments, or the scale of the specific CPP scenario.

DRL-based approaches can tackle CPP problems directly and without the need for previous decomposition but are generally underexplored in the context of autonomous UAV coverage missions. One of the rare examples is given in~\cite{dong2021collaborative}, where a Q-learning-based algorithm is proposed for coordinating multiple UAVs on a coverage mission. In contrast to the scenario we investigate, the coverage mission is set in an environment not requiring obstacle avoidance or recharge. It is not based on directly using a map as input to the DRL agent. Our previous work~\cite{theile2020uav, theile2021uav}, which focuses on generalizing learning over scenario parameters and map-processing techniques is related to this work, but only tackles the CPP problem \textit{without} recharge. 

Considering general CPP approaches for ground-based robots, several works utilize some form of DRL in their solution. In~\cite{chen2019adaptive}, the authors develop an abstraction model for CPP scenarios to find a specific coverage path solution with DRL. In~\cite{heydari2021reinforcement}, a CPP problem is reformulated as an optimal time-stopping problem, and the authors demonstrate a solution to this problem with a deep Q-network approach for indoor environments. Kyaw et al.~\cite{kyaw2020coverage} use a classical cellular decomposition approach to repose the CPP problem as a traveling salesman problem, then solve it with the help of the REINFORCE algorithm. A more practical perspective is given in~\cite{lakshmanan2020complete}, where a CPP problem for specific tetromino-shaped reconfigurable robots is solved with the help of an actor-critic type DRL algorithm and compared to various other existing CPP algorithms. In the context of a medical surface disinfecting robot, a graph-based environment representation is used in~\cite{nasirian2021efficient} to solve a CPP problem using a deep Q-network DRL approach. In~\cite{jonnarth2023end}, the authors investigate another form of the CPP problem in unknown environments and also use a map-based DRL approach, but with the combination of local sensory inputs of the robotic platform. All mentioned approaches focus on a scenario where the complete free space is the coverage target, and no recharging is needed. Several other tasks, for which DRL-based solutions were suggested, are closely CPP-related. Typical examples are patrolling~\cite{luis2020deep, piciarelli2020drone}, which is the task of checking an area of interest repeatedly, and the exploration of unknown environments~\cite{hu2020voronoi, niroui2019deep, garaffa2021reinforcement}.

With safety being a top concern in path planning for autonomous UAVs, we propose using model-based action masking to constrain the actions available to the DRL agent to a safe subset. Regarding safe learning approaches in RL, this can be classified as a straightforward certifying learning-based control approach~\cite{brunke2022}. Action masking has been suggested in other domains, e.g., in autonomous driving scenarios~\cite{krasowski2020safe} or for the vehicle routing problem~\cite{nazari2018reinforcement}. Independently of a specific application, the authors in~\cite{huang2022closer} investigate the consequences of using action masking in policy gradient DRL algorithms, while general action space shaping is investigated in~\cite{kanervisto2020action}.

\section{Problem Formulation}
\label{sec:problem}
\begin{figure}
\centering
\begin{minipage}{0.49\columnwidth}
\includegraphics[width=\textwidth]{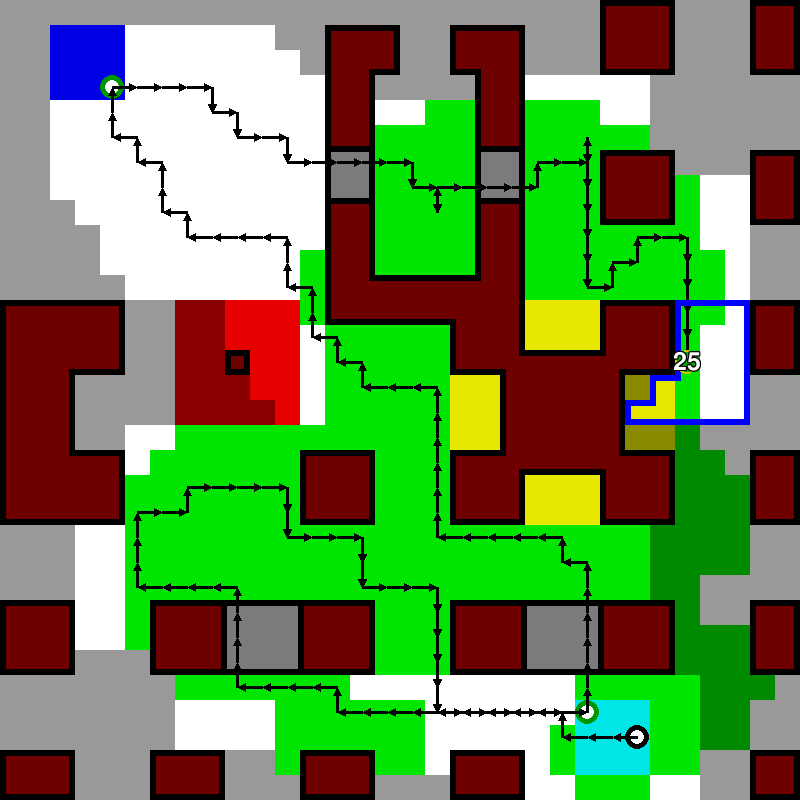}
\end{minipage}%
\hfill%
\begin{minipage}{0.50\columnwidth}
\small
\renewcommand{\arraystretch}{1.0}
\begin{tabular*}{\textwidth}{cl}
\toprule[1.5pt]
 & Description\\
\midrule
\includegraphics[align=c,height=.3cm]{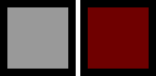} & Low/High buildings\\
\includegraphics[align=c,height=.3cm]{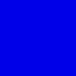} & Landing zone\\
\includegraphics[align=c,height=.3cm]{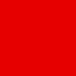} & No-fly zone (NFZ)\\
\includegraphics[align=c,height=.3cm]{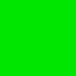} & Target zone \\
\includegraphics[align=c,height=.3cm]{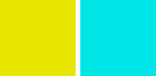} & Target + NFZ/Landing\\
\includegraphics[align=c,height=.3cm]{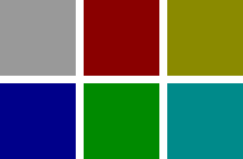} & Cells not covered\\
\includegraphics[align=c,height=.3cm]{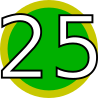} & Agent with battery\\
\includegraphics[align=c,height=.3cm]{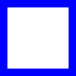} & Field of view (FoV)\\
\includegraphics[align=c,width=.5cm]{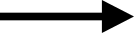} & Trajectory\\
\includegraphics[align=c,height=.3cm]{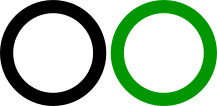} & Take-off/Charging point\\
\bottomrule[1.5pt]
\end{tabular*}
\end{minipage}
\caption{Example state of a UAV in a coverage path planning grid-world problem on the left, showing the covered area, trajectory, and field of view, with a legend on the right.}
\label{fig:cpp}
\end{figure}

This section describes the \gls*{cpp} problem by first introducing the UAV grid-world model followed by the \gls*{cpp} objective. Figure~\ref{fig:cpp} visualizes the problem in an example state.

\subsection{UAV Grid-World}

We consider a square grid-world of size $m \times m \in \mathbb{N}^2$ with cell width and height $c$ where the set of all cell indices is given by $\mathcal{M}$. Each cell in the grid-world can be a low or high obstacle, a landing zone, a \gls*{nfz}, or empty. The grid-world can thus be described through three Boolean matrices $B, L, Z \in \mathbb{B}^{m\times m}$, where the matrices indicate the presence of an obstacle, a landing zone, and \gls*{nfz} in each cell, respectively. A low obstacle that can be flown over is only defined as an obstacle, while a high obstacle that cannot be passed over is also defined as an \gls*{nfz}. Landing zones can neither be obstacles nor \glspl*{nfz}. To reduce the use of indexing in the following descriptions, we define the set of obstacles, landing zones, and \glspl*{nfz} as $\mathcal{B}$, $\mathcal{L}$, and $\mathcal{Z}$, which contain the indices of all cells for which the respective Boolean matrices are \textit{true}.

The \gls*{uav} is assumed to occupy one cell at a time described through its position $\mathbf{p}_t \in \mathcal{M}$. It is either flying at cruise altitude $h$ or landed at altitude 0, indicated through its landed state $l_t\in\mathbb{B}$, which is \textit{true} if the \gls*{uav} is landed and \textit{false} if it is flying. The \gls*{uav} has a battery with a known charge, which is defined through the remaining flying time $b_t$, indicating the number of steps that can be taken until the battery is empty.

The \gls*{uav} can move to adjacent cells, take off, land, or charge, which is described through an action $a_t \in \mathcal{A}$ where
\begin{equation}\label{eq:action_space}
\mathcal{A} = \{\textit{east}, \textit{north}, \textit{west}, \textit{south}, \textit{take off}, \textit{land}, \textit{charge}\}.
\end{equation}
The position of the \gls*{uav} evolves according to 
\begin{equation}
    \mathbf{p}_{t+1} = \mathbf{p}_{t} + f_m[a_t],
\end{equation}
where $f_m$ is the motion lookup table defined as
\begin{equation}\label{eq:motion}
      f_m = 
    \Bigg\{
    \underbrace{\begin{bmatrix}1\\0\end{bmatrix}}_{\text{east}}, 
    \underbrace{\begin{bmatrix}0\\1\end{bmatrix}}_{\text{north}},
    \underbrace{\begin{bmatrix}-1\\0\end{bmatrix}}_{\text{west}},
    \underbrace{\begin{bmatrix}0\\-1\end{bmatrix}}_{\text{south}},
    \underbrace{\begin{bmatrix}0\\0\end{bmatrix}}_{\text{take off}},
    \underbrace{\begin{bmatrix}0\\0\end{bmatrix}}_{\text{land}},
    \underbrace{\begin{bmatrix}0\\0\end{bmatrix}}_{\text{charge}}
    \Bigg\}.
\end{equation}
The landed state evolves according to
\begin{equation}
    l_{t+1} = \begin{cases}
        \textit{true}, &\text{if} ~ a_t = \textit{land}, \\
        \textit{false}, &\text{if} ~ a_t = \textit{take off}, \\
        l_t, & \text{else}.
    \end{cases}
\end{equation}
The battery state follows
\begin{equation}
    b_{t+1} = \begin{cases}
         \min\{b_{t} + b_c, b_{\text{max}}\}, \quad &\text{if}~ a_t = \textit{charge}, \\
         b_t - 1,  &\text{else},
    \end{cases}
\end{equation}
with $b_c \in \mathbb{N}$ being a constant charge amount, and $b_{\text{max}}$ as the maximum capacity of the battery.

The constraints of the \gls*{uav} grid-world problem are defined through action and state constraints. Action constraints are given as
\begin{align}
\begin{split}
         ~ l_t &\land a_t = \textit{take off} \\
    \lor ~ \phantom{\lnot} l_t &\land b_t < b_\text{max} \land a_t = \textit{charge} \\
    \lor ~ \lnot l_t &\land a_t \in \{\textit{east}, \textit{north}, \textit{west}, \textit{south}\}\\
    \lor ~ \lnot l_t &\land \mathbf{p}_t \in \mathcal{L} \land a_t = \textit{land}.
\end{split}\label{eq:action_constraints}
\end{align}
The \gls{uav} can only take off if landed, can only charge if landed and the battery is not full, can only move if flying, and can only land if currently flying and in a landing zone. Constraints on the state are summarized as 
\begin{subequations}\label{eq:state_constraints}
    \begin{align}
        \mathbf{p}_t &\notin \mathcal{Z}, \label{eq:const_nfz} \\
        b_t &> 0 ~\lor~ l_t, \label{eq:const_bat}
    \end{align}
\end{subequations}
which define that the \gls{uav} cannot be in an \gls{nfz}, and the battery cannot be empty unless landed. 

\subsection{Coverage Path Planning}
The \gls{cpp} problem extends the UAV grid-world problem by defining target zones summarized through $C_t \in \mathbb{B}^{m\times m}$, with $\mathcal{C}_t$ being the set of indices of cells defined as targets. The target evolves according to
\begin{equation}
    \mathcal{C}_{t+1} = \{\forall \mathbf{x} \in \mathcal{C}_t ~|~ \lnot v(\mathbf{p}_{t+1}, \mathbf{x}, B) \}
\end{equation}
in which $v:\mathcal{M}\times\mathcal{M}\times \mathbb{B}^{m\times m} \mapsto \mathbb{B}$ indicates whether cell~$\mathbf{x}$ is visible from the next position~$\mathbf{p}_{t+1}$ given the obstacles~$B$. It thus defines the current \gls{fov} of the \gls{uav}. 
The initial state of the system is set as
\begin{subequations}
    \begin{align}
        \mathbf{p}_0 &\in \mathcal{L}, \label{eq:init_pos} \\
        b_0 &\in [\beta b_{\text{max}}, b_{\text{max}}], \label{eq:init_bat}
    \end{align}
\end{subequations}
with $\beta\in (0,1]$. The initial coverage target $\mathcal{C}_0$ consists of multiple random patches with $\mathcal{C}_0 \cap \mathcal{B} = \{\}$, i.e., no target zones in cells occupied by obstacles. However, target zones can be in NFZs, which need to be covered by flying adjacent to the cells. The initial landing state depends on the objective. We formulate two objectives for the \gls{cpp} problem.

\subsubsection{Power-constrained CPP without recharge}
In the power-constrained \gls{cpp} problem without recharge the initial landing state of the \gls{uav} is $l_0 = \textit{false}$ and the objective is
\begin{align}
    \begin{split}
        & \min |\mathcal{C}_{T}| \\
        \text{s.t.}\quad &\eqref{eq:action_constraints}  \land \eqref{eq:const_nfz} \land \eqref{eq:const_bat} , ~\forall t\in[0, T]\\
        \land ~ & \eqref{eq:init_pos} \land \eqref{eq:init_bat} \land l_0 = \textit{false},
    \end{split}
\end{align}
in which
\begin{equation}
        T = \operatorname{arg}\min_t t \qquad
        \text{s.t.}\quad  l_t.
\end{equation}
The objective is to minimize the remaining target zone until the agent lands for the first time, while adhering to the action and state constraints, with the initial state flying over a landing zone with a randomized battery level. This objective was used in previous work~\cite{theile2020uav, theile2021uav}.

\subsubsection{Power-constrained CPP with recharge}
In the power-constrained \gls{cpp} problem with recharge the initial landing state is $l_0=\textit{true}$ and the objective is
\begin{align}
    \begin{split}
        & \min T \\
        \text{s.t.}\quad & \mathcal{C}_T = \{\} \land ~ l_T \\
        \land ~&\eqref{eq:action_constraints} \land \eqref{eq:const_nfz} \land \eqref{eq:const_bat} , ~\forall t\in[0, T] \\
        \land ~ & \eqref{eq:init_pos} \land \eqref{eq:init_bat} \land l_0 = \textit{true}.
    \end{split}\label{eq:obj_recharge}
\end{align}
The objective is to minimize the number of steps until the remaining coverage target is empty and the \gls{uav} is landed, while adhering to the action and state constraints, with the initial state landed in a landing zone with a randomized battery level. This objective is closer to a realistic UAV CPP scenario, but also much more challenging than the previous one as it extends over a significantly longer time-horizon. Therefore, this paper focuses on solving this power-constrained \gls{cpp} problem \textit{with} recharge.

\section{Methodology}
\label{sec:methodology}
We solve the described problem using \gls{rl} with model-based action masking for constraint satisfaction. To solve the problem with \gls{rl}, it needs to be formulated as a \glsentryfull{mdp}. Since we utilize specific observation functions, we describe the problem as a \gls{pomdp}, which is defined by the tuple $(\mathcal{S}, \mathcal{A}, R, P,\Omega, O, \gamma)$. It contains the set of possible states $\mathcal{S}$, the set of possible actions $\mathcal{A}$, the reward function $R: \mathcal{S} \times \mathcal{A} \times \mathcal{S} \mapsto \mathbb{R}$, and the deterministic state transition function $P: \mathcal{S} \times \mathcal{A} \mapsto S$. Furthermore, an observation space $\Omega$ describes the agent's observations, and the observation function $O:\mathcal{S}\mapsto \Omega$ defines the mapping from states to observations. The discount factor $\gamma \in [0,1]$ balances the importance of immediate and future rewards. The state space of the system can be described through
\begin{equation}\label{eq:state_space}
    \mathcal{S} = 
    \underbrace{\mathbb{B}^{m\times m \times 3}}_{\substack{\text{Environment}\\ \text{Map}}}\times 
    \underbrace{\mathbb{B}^{m\times m}}_{\substack{\text{Target}\\ \text{Map}}}\times 
    \underbrace{\mathbb{N}^2}_{\text{Position}}\times
    \underbrace{\mathbb{N}}_{\substack{\text{Battery}\\ \text{Level}}}\times
    \underbrace{\mathbb{B}}_{{\text{Landed}}}.
\end{equation}
In principle, the discrete action space is as given in~\eqref{eq:action_space}. However, we mask out infeasible actions based on a safety model.

\subsection{Action Masking through Safety Modeling}
We consider three different levels of action space restrictions: \textit{valid}, \textit{immediate}, and \textit{invariant}. The valid action mask is defined based on actions that violate the action constraints in~\eqref{eq:action_constraints}. These actions are considered invalid as their outcomes are undefined, such as taking off if already flying or trying to fly east while landed. It can be defined as:
\begin{equation}
\mathcal{A}^+_\text{val}(s_t) = \{\forall a \in \mathcal{A} ~|~ \eqref{eq:action_constraints} \},\label{eq:invalid}
\end{equation}
with $s_t\in\mathcal{S}$. The valid action mask is the most commonly used in the literature~\cite{kanervisto2020action}.

The immediate action mask disallows actions leading to an immediate next state that violates any constraint, i.e., an action after which the UAV is in an \gls{nfz}. Based on the constraints~\eqref{eq:action_constraints} and~\eqref{eq:const_nfz}, it can be defined as:
\begin{equation}
\mathcal{A}^+_\text{imm}(s_t) = \{\forall a \in \mathcal{A}^+_\text{val}(s_t) ~|~ \mathbf{p}_t + f_m[a] \notin \mathcal{Z} \}.\label{eq:immediate}
\end{equation}

Finally, the invariant action mask is intended to disallow actions that lead to unrecoverable states. In the grid-world, a state is unrecoverable if the battery level is insufficient to return to a landing zone and recharge, i.e., state constraint~\eqref{eq:const_bat}. The invariant action mask can be defined by first computing the distance from the nearest landing zone for each cell, using the distance matrix $D\in \mathbb{N}^{m\times m}$ defined recursively as:
\begin{equation}
D_{i,j} =
\begin{cases}
1, & \text{if}~ L_{i,j} \\
\infty, & \text{if}~ Z_{i,j} \\
1 + \min\{D_{i+1,j}, D_{i-1,j},~ & \\ \hfill D_{i,j+1}, D_{i,j-1}\}, & \text{else}.
\end{cases}
\end{equation}
The recursion stops at landing zones and NFZs, setting the distance to 1 and $\infty$, respectively. Using the function $d_L:\mathcal{M}\mapsto\mathbb{N}$ that indexes $D$, the state-dependent invariant action space can then be defined as:
\begin{equation}
\label{eq:invariant}
\mathcal{A}^+_\text{inv}(s_t) = \{\forall a \in \mathcal{A}^+_\text{imm}(s_t) ~|~ d_L(\mathbf{p}_t + f_m[a]) \leq b_t - 1 \}.
\end{equation}
All immediately safe actions that lead to a state that is at maximum $b_t - 1$ steps away from a landing zone are invariantly safe. The mask is applied by setting the infeasible action logits of the actor to $-\infty$, which is further described in Section~\ref{sec:nn}. Section~\ref{sec:mask_results} ablates the usage of different mask levels.

\subsection{Reward Function and Discount Factor}
The ideal reward function that reflects the objective in~\eqref{eq:obj_recharge} only contains a penalty $-r_m$ for each step taken in the environment since, with invariant action masking, the episode only terminates when the remaining coverage target is empty and the agent is landed. However, since the agent learns from scratch, it is highly unlikely, effectively impossible, to solve the problem randomly to observe the termination condition even once. Therefore, some reward shaping is necessary such that the agent is guided toward solving the problem first and then minimizing the number of required steps. We chose to give a reward $r_c$ for every covered cell at the moment of coverage, leading to the following reward function:
\begin{equation}
    r_t = r_c (|\mathcal{C}_{t}| - |\mathcal{C}_{t+1}|) - r_m.
\end{equation}
With a discount factor $\gamma=1$, this reward function still yields a return $\mathrm{R}= \sum_{t=0}^T \gamma^t r_t = r_c |\mathcal{C}_{0}| - r_m T$, which is maximized only by decreasing $T$. However, as before, the agent initially does not know how to solve the problem and, thus, the value estimate would be unstable, potentially tending towards $\pm\infty$ depending if $r_c$ or $r_m$ dominates the estimate. On the other hand, a discount factor $\gamma < 1$ often results in greedy behavior as early coverage becomes more important than finishing the coverage mission overall faster. To avoid this conundrum, we start with a lower discount factor allowing the agent to learn how to successfully complete the mission and for the value estimate to stabilize. Throughout training, we gradually increase the discount factor $\gamma \rightarrow 1$, specifically we use 
\begin{equation}\label{eq:discount}
    \gamma = 1- (1- \gamma_0) \gamma_r^{\,s/\gamma_s},
\end{equation}
with the starting value $\gamma_0$, decay rate $\gamma_r$, decay steps $\gamma_s$, and the current interaction step $s$. 

The suggested discount factor scheduling proves advantageous within finite-horizon problems where the horizon is contingent on the agent's performance, initially creating the illusion of an infinite problem. This approach stabilizes initial value estimates while facilitating long-horizon optimization during the advanced phases of training. Section~\ref{sec:discount_results} ablates different discount factor scheduling parameters.

\subsection{Position History} \label{subsec:position_history}
\begin{figure}
    \centering
    \href[pdfnewwindow]{https://youtu.be/BeKn-VVhrz0?t=3}{
    \begin{subfigure}[t]{0.48\columnwidth}
        \centering
        \includegraphics[width=\textwidth]{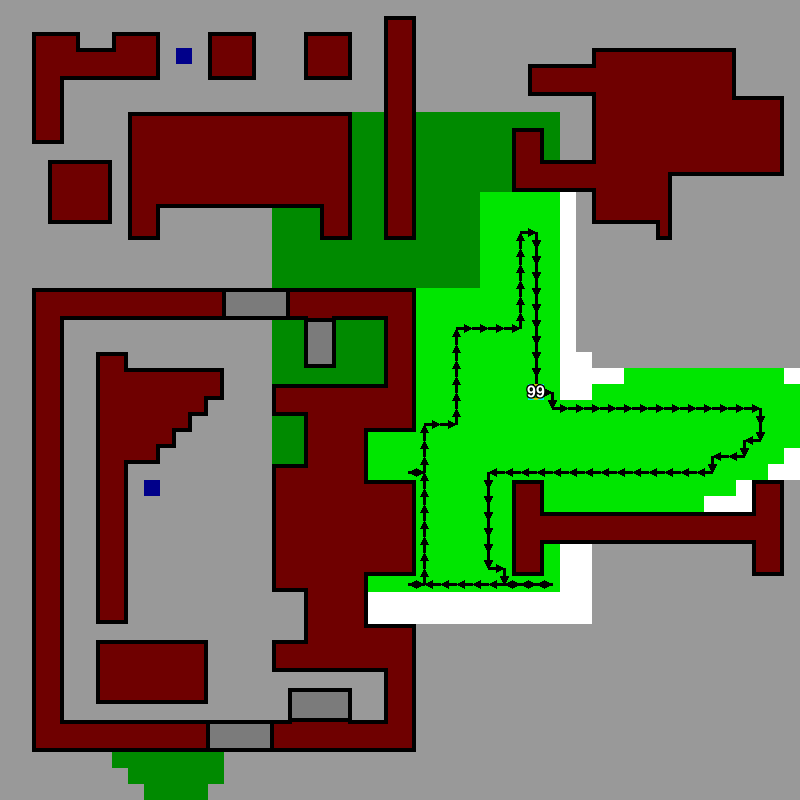}
        \caption{The agent is stuck in the loop of taking off, landing, and recharging one step.}
    \end{subfigure}\hfill%
    \begin{subfigure}[t]{0.48\columnwidth}
        \centering
        \includegraphics[width=\textwidth]{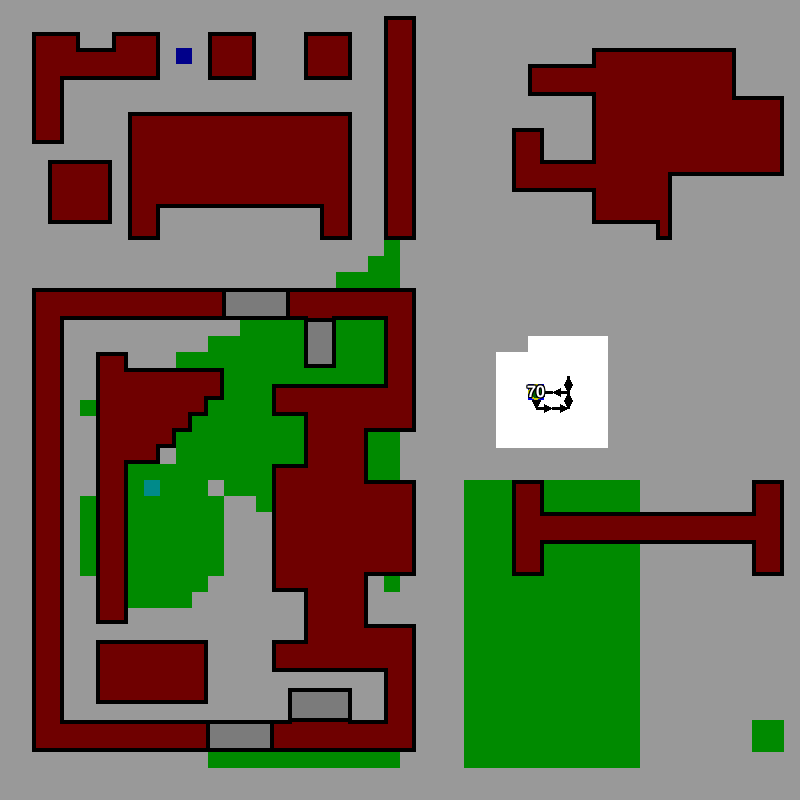}
        \caption{The agent goes back and forth until the safety masking forces it to fly back.}
    \end{subfigure}
    }
    \caption{Two scenarios in which the agent is stuck in infinite loops. By clicking on the images, a link to a video can be opened that shows the behavior.}
    \label{fig:loops}
\end{figure}

An added challenge through the recharge ability is that the agent can recreate the same state. If the agent takes off, flies around without covering any new target, and then returns to the same landing zone cell to recharge, the same state will be observed after charging. If the actions chosen by the agent are deterministic, the agent will be stuck in this loop, indefinitely. Figure~\ref{fig:loops} shows two examples of this behavior.

For the agent to avoid getting stuck in such a loop, it either needs to act stochastically or the associated observation needs to change between loops. To alter the observation and help the agent to learn to avoid repeating the same behavior, we augment the state with a position history layer $H \in \mathbb{R}^{m\times m}$, which evolves according to
\begin{equation}
    H_{t+1,i,j} = \begin{cases}
        1, &\text{if}~ \mathbf{p}_{t+1} = (i,j), \\
        \alpha H_{t,i,j}, & \text{else},
    \end{cases}
\end{equation}
with $H_0$ being 1 at $\mathbf{p}_0$ and 0 otherwise. All values except the current position in the position history decay with \mbox{$\alpha < 1$}. The position history is concatenated with the other map layers in the observation. Note that we do not add any reward shaping to avoid the revisitation of cells. Section~\ref{sec:pos_hist_results} ablates the usage of the position history.

\subsection{Global-Local Observation}
We use the global-local observation method from previous work~\cite{theile2021uav} as the observation function. In this method, the agent observes the environment as two maps centered around the agent, a global compressed map and a local full-resolution map. Centering the map was shown to be essential for generalization capabilities in~\cite{bayerlein2020uav}. The motivation for using the global-local observation function is that it outperformed centered observations in~\cite{theile2021uav} and in initial experiments for this paper. The global-local preprocessing appears to aid the agent significantly, and additionally it greatly reduces memory requirements for storing experiences compared to the centered map.

The observation processing pipeline is visualized on the left side of Figure~\ref{fig:nn_arch}. It starts by centering the map around the agent's position and padding it with obstacles and \glspl{nfz}, increasing the size to $m_c = 2m -1$. The global map downsamples the centered map by the global map scaling factor $g$, and the local map crops out an $l \times l$ central region, with the local map size $l$. The resulting observation space $\Omega$, is given as
\begin{equation*}
    \Omega = \Omega_g\times \Omega_l\times
    \mathbb{R}^2_{\phantom{l}},
\end{equation*}
containing the global map
$
    \Omega_g = \mathbb{R}^{\lceil\frac{m_c}{g}\rceil\times \lceil\frac{m_c}{g}\rceil\times 5},
$
the local map
$
    \Omega_l = \mathbb{R}^{l\times l\times 5},
$
and the state scalars, consisting of the normalized battery $b_t/b_{\text{max}}$ and the landed state $l_t$. The layers of the maps are landing zones, \glspl{nfz}, obstacles, remaining targets, and position history. For simplicity, all channels and scalars are cast to real values. 

\subsection{Neural Network Architecture}
\label{sec:nn}

\begin{figure}
    \centering
    \includegraphics[width=\columnwidth]{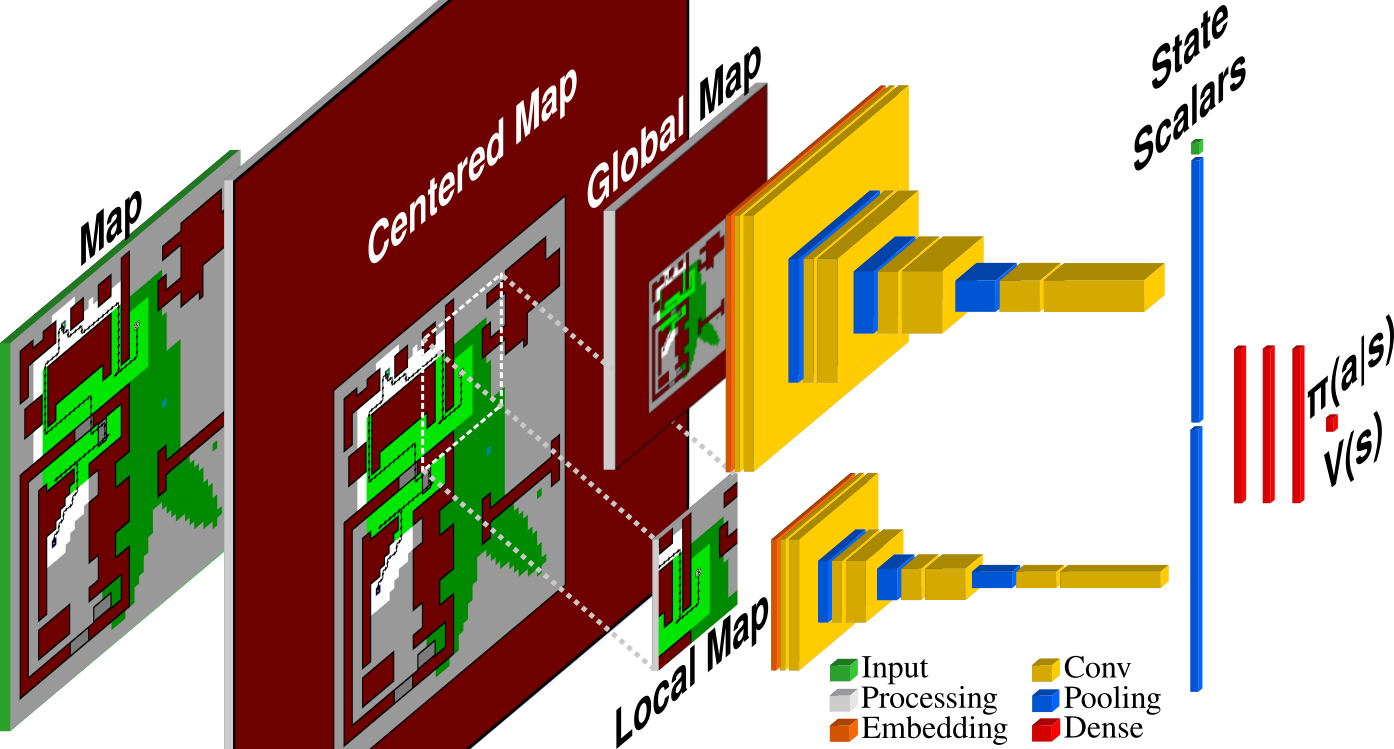}
    \caption{Map-based processing pipeline and neural network architecture with to-scale relative spatial dimensions.}
    \label{fig:nn_arch}
\end{figure}
We train the agents using the \glsentryfull{ppo} algorithm~\cite{schulman2017proximal}. PPO is a popular and effective reinforcement learning method widely used in various applications. It is an on-policy actor-critic algorithm. In our work, the parameters of the actor and critic, $\theta$ and $\phi$, parameterize two neural networks with the same structure, shown in Figure~\ref{fig:nn_arch}. After creating the global and local maps, they are embedded by a $1\times 1$ linear convolution layer into 32 channels. After that, both maps are processed similarly with two $3\times 3$ convolution layers with ReLU activation function, where the second layer doubles the number of channels, followed by a $2\times 2$ max pooling. This is repeated three more times, after which the spatial dimensions are reduced by a last max pooling, yielding two 512-element vectors. These vectors are concatenated with the state scalars input and processed through three hidden layers of size 256 and with ReLU activation function. The output layer has no activation function and is a single neuron for the critic and $|\mathcal{A}|$ neurons for the actor. At this point, action masking can be applied by setting the output values of the actor to $-\infty$ for all infeasible actions. The policy $\pi_\theta(a|s)$ is a $\operatorname{softmax}$ of the masked output values. 

\section{Evaluation Setup}
\label{sec:simulation}

\subsection{Heuristic for Comparison}
As the methods described in the literature are not directly applicable to the scenario in this paper, we developed a model-based greedy heuristic as a baseline for comparison. The heuristic determines the nearest reachable cell from which a coverage target cell is visible and flies towards it. If no reachable cell exists, it lands to recharge fully. Algorithm~\ref{alg:heuristic} provides a pseudocode of the heuristic. It uses the distance functions $d:\mathcal{M}\times\mathcal{M}\mapsto \mathbb{N}$ that gives the distance in steps from one cell to another, and $d_L:\mathcal{M}\mapsto \mathbb{N}$ that provides the distance of a given cell to the closest landing zone. Lines 1-3 state that whenever the agent is landed, it should fully recharge and then take off. If there is no remaining target zone, the agent should fly to the nearest landing zone and land to complete the task (lines 4-7). Otherwise, line 8 computes all cells of interest that are not \glspl{nfz} and from which a target cell is in \gls{fov}. Using these, line 9 determines all reachable cells of interest where the agent could fly and return to a landing zone within its battery constraint. If there is no reachable cell of interest, lines 10-14 compute all reachable landing zones to select the one closest to a cell of interest to fly there and land. This step is essential if the nearest cell of interest is more than a full charge away. If a reachable cell of interest exists, the agent should select the closest and fly toward it (lines 15-16).

\begin{algorithm}[t]
\caption{Greedy CPP Heuristic}\label{alg:heuristic}
\KwData{$\mathbf{p}_t, b_t, l_t, \mathcal{C}_t, \mathcal{L}, \mathcal{Z}, \mathcal{B}, D$}
\KwResult{Plan}
\If{$l_t$}{
\Return Recharge fully and take off
}
\If{$\mathcal{C}_t$ is empty}{
    $ \mathbf{x} \gets \operatorname{arg}\min_{ \mathbf{l} \in\mathcal{L}} d(\mathbf{p}_t, \mathbf{l})$ \\
    \Return Fly to $\mathbf{x}$ and land
}
$\mathcal{I} \gets \{ \forall \mathbf{p} \in \mathcal{M} ~|~ \mathbf{p} \notin \mathcal{Z} \land  v(\mathbf{p}, \mathbf{x}, B), ~\exists \mathbf{x} \in \mathcal{C}_t\}$ \\
$\mathcal{I}_r \gets \{ \forall \mathbf{y} \in \mathcal{I} ~|~ d_L(\mathbf{y}) + d(\mathbf{p}_t, \mathbf{y}) < b_t \}$ \\
\If{$\mathcal{I}_r$ is empty}{
    $\mathcal{L}_r \gets \{ \forall \mathbf{l} \in \mathcal{L} ~|~ d(\mathbf{p}_t, \mathbf{l}) < b_t \}$ \\
    $\mathbf{x} \gets \operatorname{arg}\min_{\mathbf{l}\in\mathcal{L}_r} ( d(\mathbf{l}, \mathbf{y}),~\exists \mathbf{y}\in\mathcal{I})$ \\
    \Return Fly to $\mathbf{x}$ and land
}
$\mathbf{x} \gets \operatorname{arg}\min_{\mathbf{y}\in\mathcal{I}_r} d(\mathbf{p}_t, \mathbf{y})$ \\
\Return Fly to $\mathbf{x}$
\end{algorithm}

In the experiments, the heuristic is used to give a baseline against which the performance of the learned agents is measured. We use the relative percentage deviation (RPD) measure to evaluate the performance in the required steps to solve the \gls{cpp} problem. It is computed as 
\begin{equation}\label{eq:rpd}
    \text{RPD} = \frac{\text{steps}_\text{agent} - \text{steps}_\text{heuristic}}{\text{steps}_\text{heuristic}},
\end{equation}
with $\text{steps}_\text{agent}$ and $\text{steps}_\text{heuristic}$ being the steps needed to solve the problem by the agent and heuristic, respectively.

\subsection{Agents and Maps}
\begin{figure}
    \centering
    \begin{subfigure}[t]{0.19\columnwidth}
        \centering
        \begin{tikzpicture}[baseline={(text.base)}]
            \node (text) {\textbf{Manhat.\vphantom{p}}};
            \node[rectangle,draw=black,line width=1pt,inner sep=0pt,below=0mm of text] (img) {\includegraphics[width=0.88\textwidth]{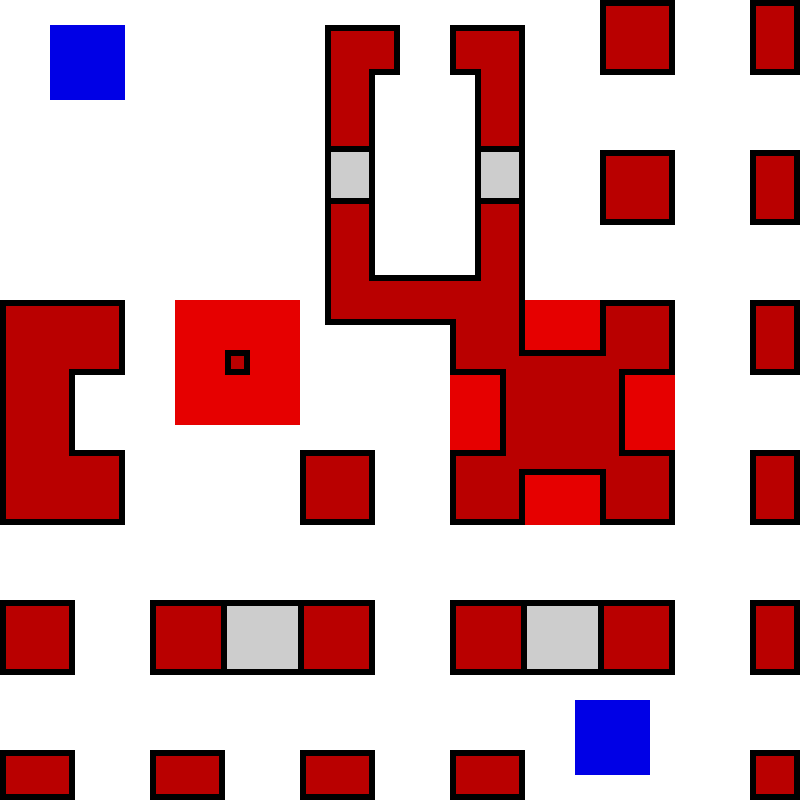}};
        \end{tikzpicture}
    \end{subfigure}%
    \begin{subfigure}[t]{0.19\columnwidth}
        \centering
        \begin{tikzpicture}[baseline={(text.base)}]
            \node (text) {\textbf{Manhat.2\vphantom{p}}};
            \node[rectangle,draw=black,line width=1pt,inner sep=0pt,below=0mm of text] (img) {\includegraphics[width=0.88\textwidth]{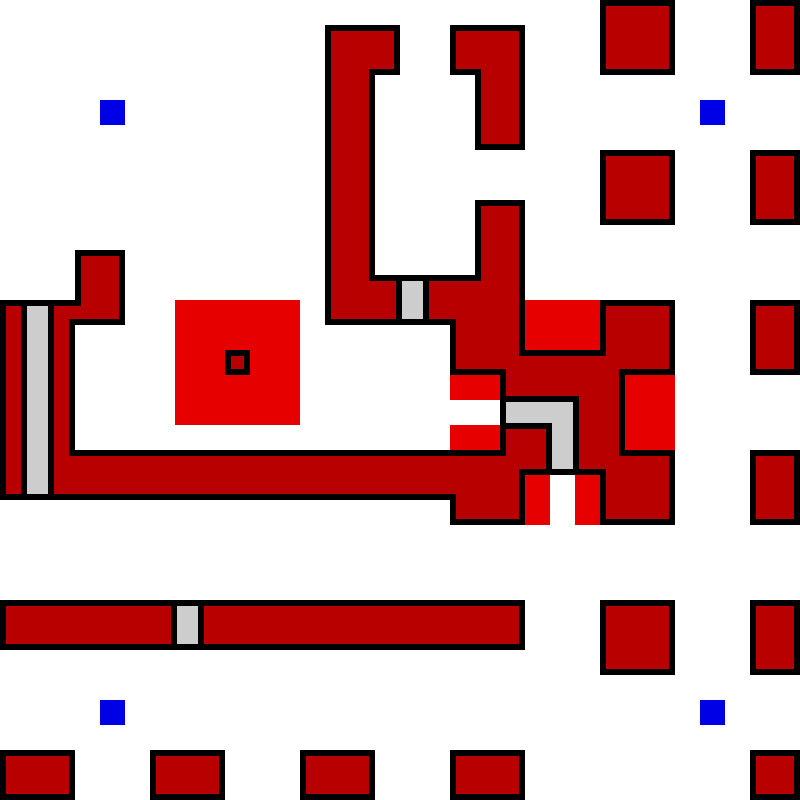}};
        \end{tikzpicture}
    \end{subfigure}%
    \begin{subfigure}[t]{0.19\columnwidth}
        \centering
        \begin{tikzpicture}[baseline={(text.base)}]
            \node (text) {\textbf{Town\vphantom{p}}};
            \node[rectangle,draw=black,line width=1pt,inner sep=0pt,below=0mm of text] (img) {\includegraphics[width=0.88\textwidth]{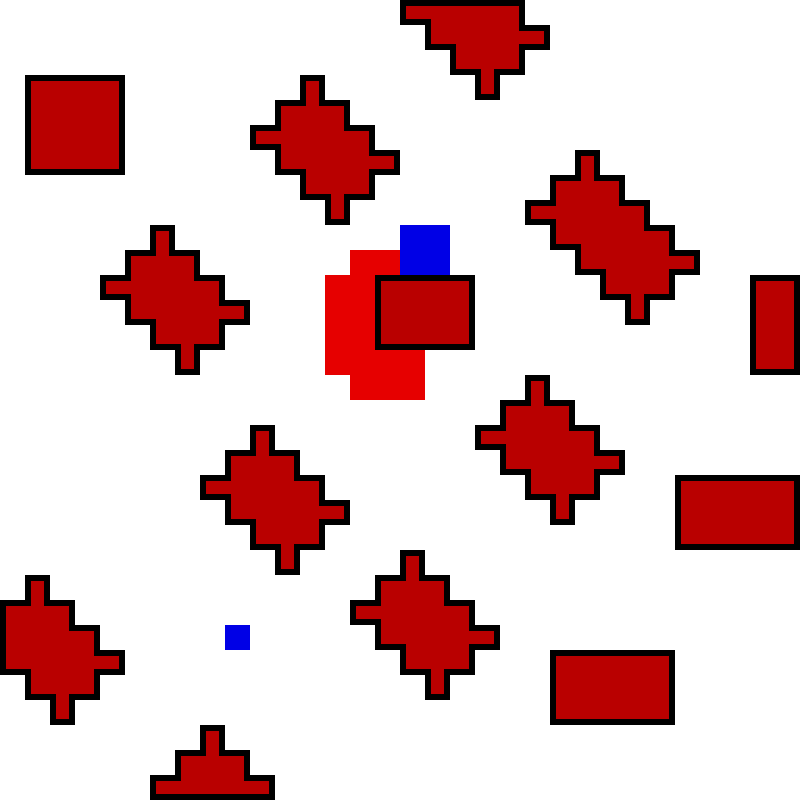}};
        \end{tikzpicture}
    \end{subfigure}%
    \begin{subfigure}[t]{0.21\columnwidth}
        \centering
        \begin{tikzpicture}[baseline={(text.base)}]
            \node (text) {\textbf{Suburban\vphantom{p}}};
            \node[rectangle,draw=black,line width=1pt,inner sep=0pt,below=0mm of text] (img) {\includegraphics[width=0.95\textwidth]{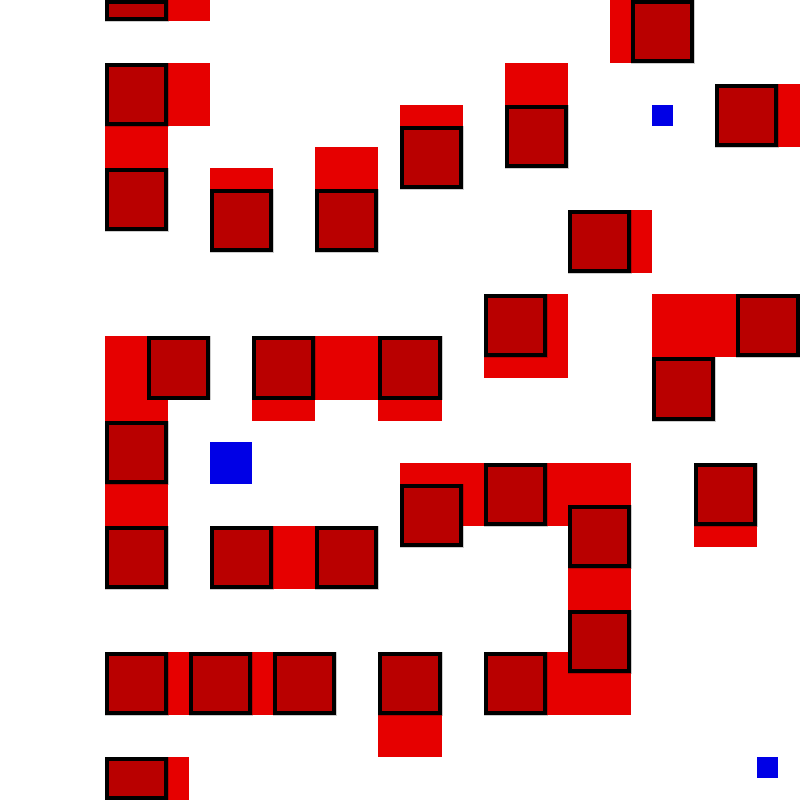}};
        \end{tikzpicture}
    \end{subfigure}%
    \begin{subfigure}[t]{0.22\columnwidth}
        \centering
        \begin{tikzpicture}[baseline={(text.base)}]
            \node (text) {\textbf{City\vphantom{p}}};
            \node[rectangle,draw=black,line width=1pt,inner sep=0pt,below=0mm of text] (img) {\includegraphics[width=0.95\textwidth]{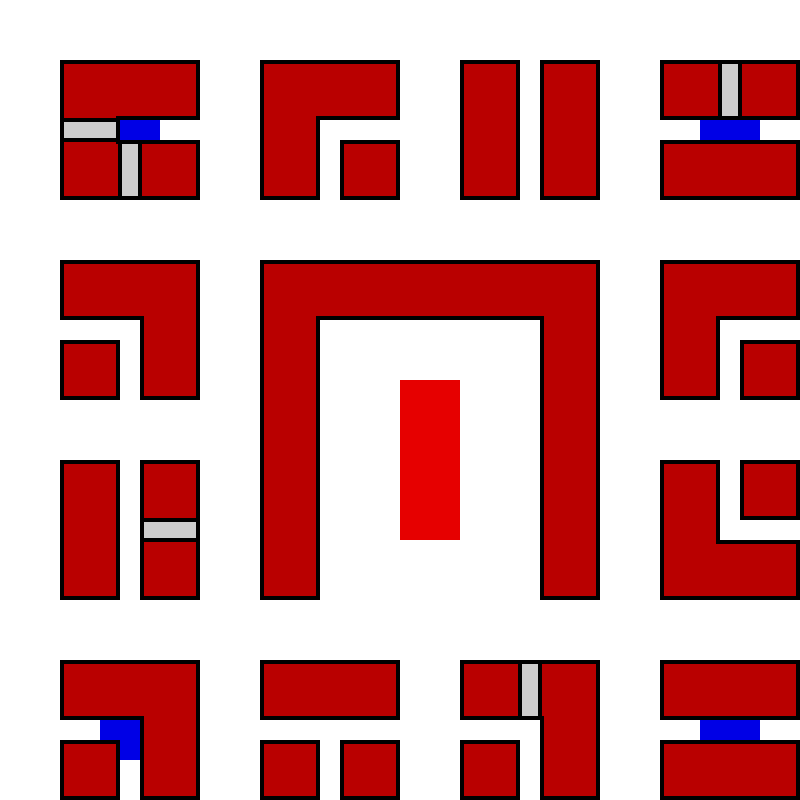}};
        \end{tikzpicture}
    \end{subfigure}
    \begin{subfigure}{0.225\columnwidth}
        \centering
        \begin{tikzpicture}[baseline={(text.base)}]
            \node (text) {\textbf{Cal\vphantom{p}}};
            \node[rectangle,draw=black,line width=1pt,inner sep=0pt,below=0mm of text] (img) {\includegraphics[width=0.98\textwidth]{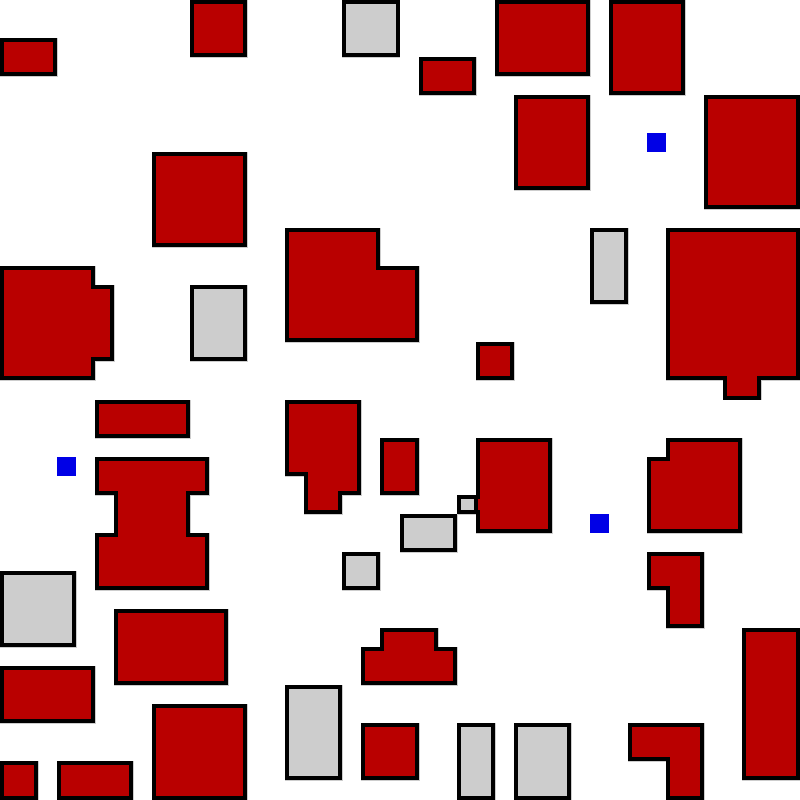}};
        \end{tikzpicture}
    \end{subfigure}%
    \begin{subfigure}{0.235\columnwidth}
        \centering
        \begin{tikzpicture}[baseline={(text.base)}]
            \node (text) {\textbf{Open}};
            \node[rectangle,draw=black,line width=1pt,inner sep=0pt,below=0mm of text] (img) {\includegraphics[width=0.98\textwidth]{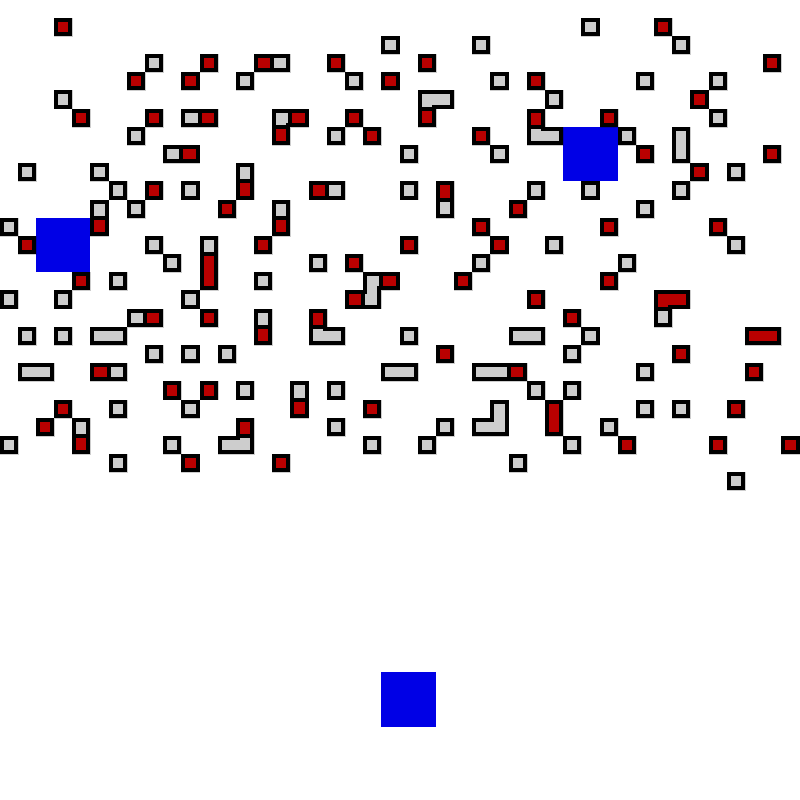}};
        \end{tikzpicture}
    \end{subfigure}%
    \begin{subfigure}{0.27\columnwidth}
        \centering
        \begin{tikzpicture}[baseline={(text.base)}]
            \node (text) {\textbf{Urban\vphantom{p}}};
            \node[rectangle,draw=black,line width=1pt,inner sep=0pt,below=0mm of text] (img) {\includegraphics[width=0.97\textwidth]{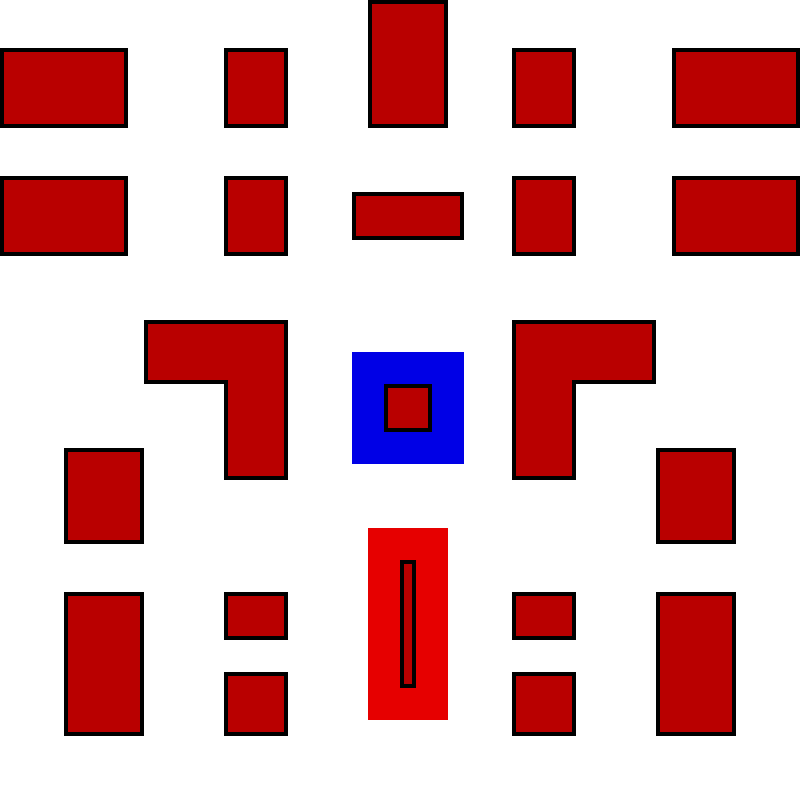}};
        \end{tikzpicture}
    \end{subfigure}%
    \begin{subfigure}{0.27\columnwidth}
        \centering
        \begin{tikzpicture}[baseline={(text.base)}]
            \node (text) {\textbf{Urban2\vphantom{p}}};
            \node[rectangle,draw=black,line width=1pt,inner sep=0pt,below=0mm of text] (img) {\includegraphics[width=0.97\textwidth]{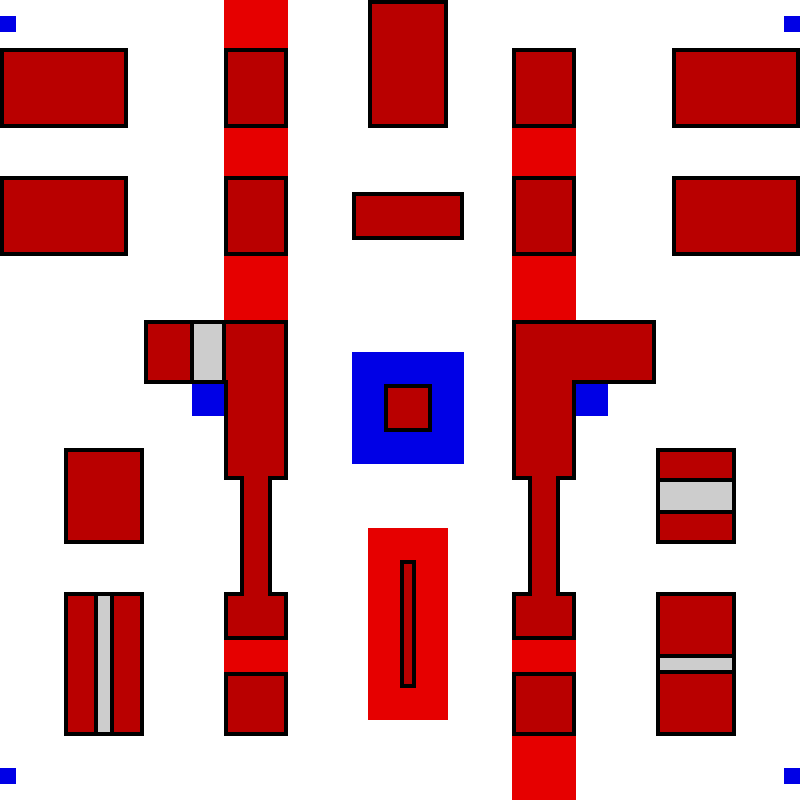}};
        \end{tikzpicture}
    \end{subfigure}
    \begin{subfigure}{0.27\columnwidth}
        \centering
        \begin{tikzpicture}[baseline={(text.base)}]
            \node (text) {\textbf{Castle\vphantom{p}}};
            \node[rectangle,draw=black,line width=1pt,inner sep=0pt,below=0mm of text] (img) {\includegraphics[width=0.97\textwidth]{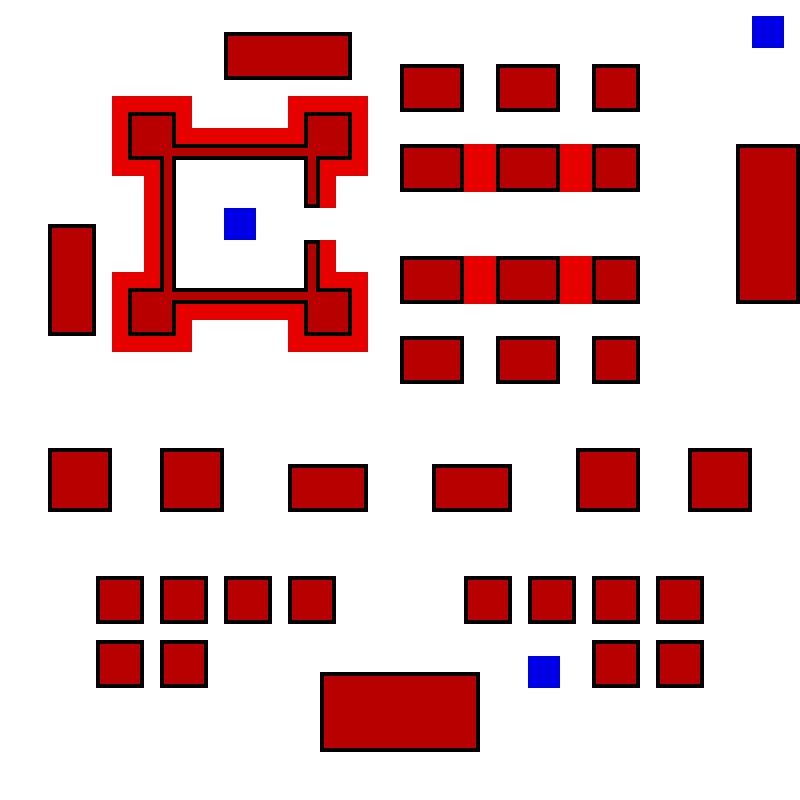}};
        \end{tikzpicture}
    \end{subfigure}%
    \begin{subfigure}{0.27\columnwidth}
        \centering
        \begin{tikzpicture}[baseline={(text.base)}]
            \node (text) {\textbf{TUM\vphantom{p}}};
            \node[rectangle,draw=black,line width=1pt,inner sep=0pt,below=0mm of text] (img) {\includegraphics[width=0.97\textwidth]{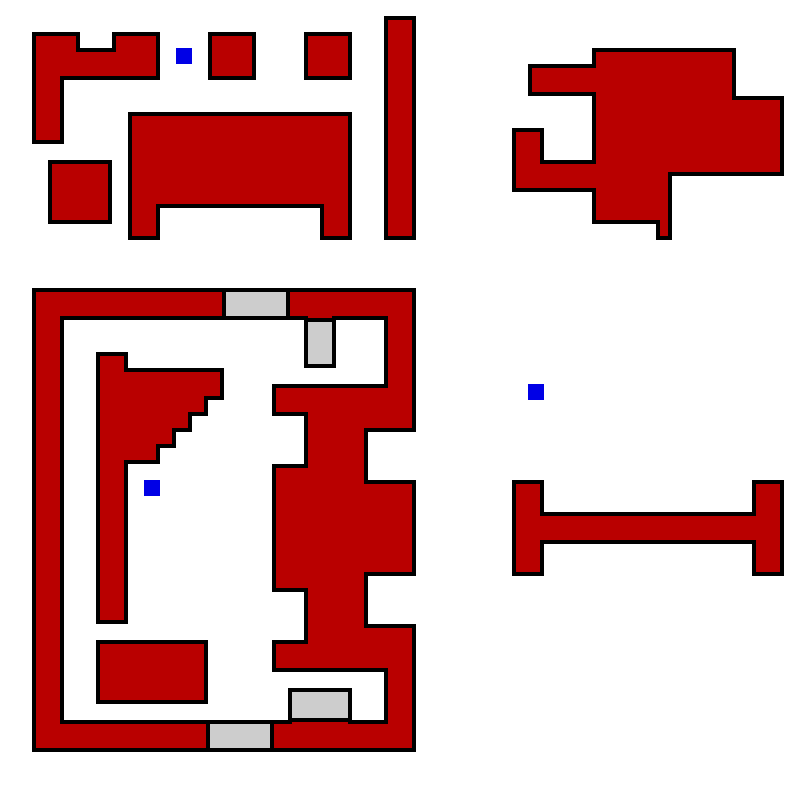}};
        \end{tikzpicture}
    \end{subfigure}%
    \begin{subfigure}{0.27\columnwidth}
        \centering
        \begin{tikzpicture}[baseline={(text.base)}]
            \node (text) {\textbf{Border\vphantom{p}}};
            \node[rectangle,draw=black,line width=1pt,inner sep=0pt,below=0mm of text] (img) {\includegraphics[width=0.97\textwidth]{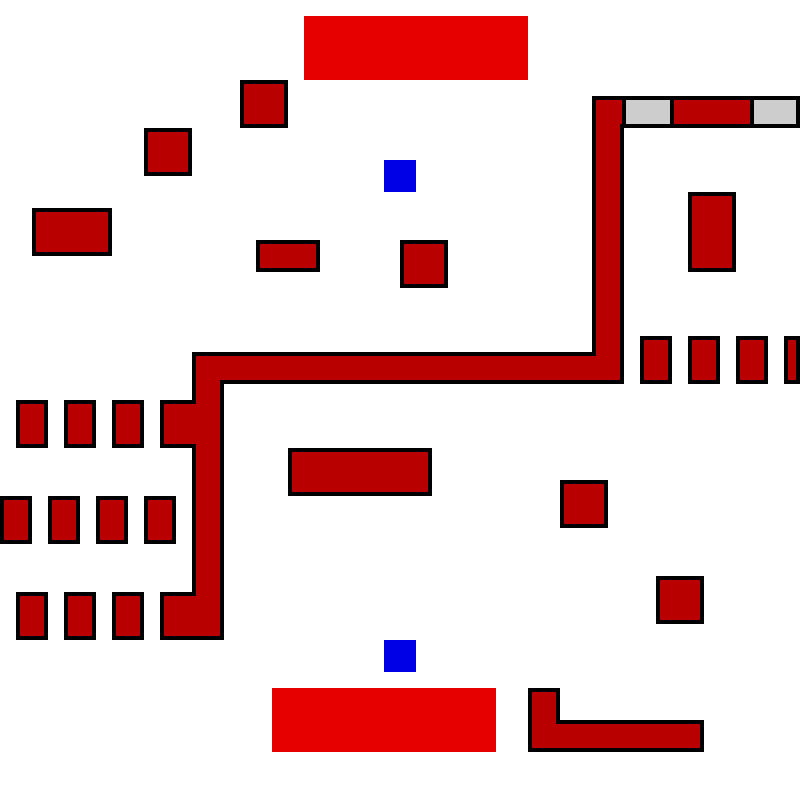}};
        \end{tikzpicture}
    \end{subfigure}%
    \caption{All maps listed in Table~\ref{tab:maps}, sorted by size.}
    \label{fig:all_maps}
\end{figure}
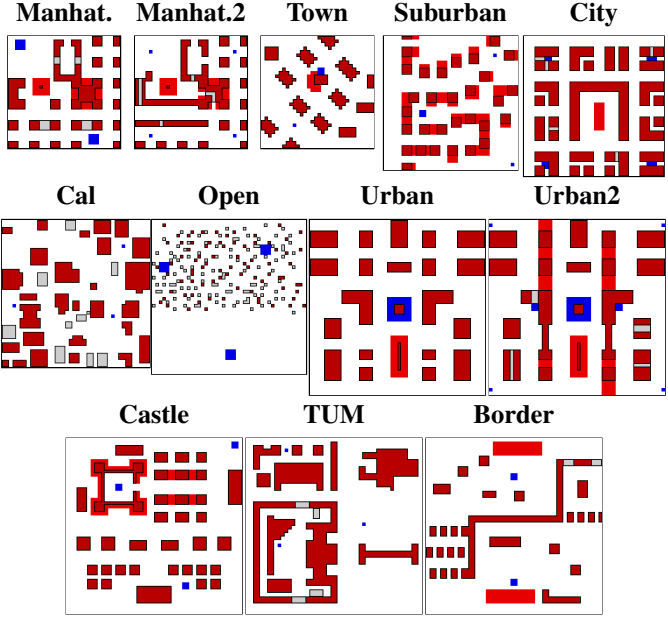

For this paper we created 12 maps, shown in Figure~\ref{fig:all_maps}, with sizes ranging from $32\times 32$ to $50\times 50$. The Manhattan map is used for illustration purposes, as it is small enough but still contains all possible CPP challenges. In the evaluation, we focus on the Suburban, Castle, TUM, Cal, and Border maps. As target zones can also be within \glspl{nfz}, the Suburban map has the specific challenge that the agent might need to pass on both sides of the wide \glspl{nfz} to cover target zones. The Castle map is challenging because of the long distances traveled to access points on the map. The castle-like shape on the top left that surrounds a landing zone further complicates the path planning. The specific challenge of the TUM map is that the agent needs to learn the difference between low and high obstacles to fly into the inner area of the bottom left region. No specific added challenges are present in the Cal map, but it contains irregular low and high buildings. Finally, the Border map is dominated by a barrier-like obstacle that divides the task area in the middle and can only be circumvented near the edges of the map. The Cal and Border maps are used for our zero-shot out-of-distribution generalization experiments to evaluate how the agent can solve simple and complicated maps that it did not encounter before.

We trained multiple agents on different subsets of maps, shown in Table~\ref{tab:maps}. The \textit{Multi} agents are trained on multiple maps, where in each training episode, a map is chosen uniformly at random. The other agents are trained only on one map, and are referred to as \textit{specialized} agents. Unless stated otherwise, all agents are trained using the parameters in Table~\ref{tab:parameters}. For each configuration, three agents were trained. Each training procedure ran for around four days on an NVIDIA A100 GPU.

\begin{table}
    \centering
    \footnotesize
    \setlength{\tabcolsep}{1pt}
    \renewcommand{\arraystretch}{1.0}
\begin{tabular}{c|c|c|c|c|c|c|c||c|c|c||c|c}
     & \rotatebox{90}{Manhattan} & \rotatebox{90}{Manhattan2} & \rotatebox{90}{Town} &  \rotatebox{90}{City} & \rotatebox{90}{Open} & \rotatebox{90}{Urban} & \rotatebox{90}{Urban2} & \rotatebox{90}{Suburban} & \rotatebox{90}{Castle} & \rotatebox{90}{TUM} & \rotatebox{90}{Cal} & \rotatebox{90}{Border} \\
     \hline
  Size & 32 & 32 & 32 & 40 & 44 & 50 & 50 & 38 & 50 & 50 & 42 & 50 \\
  T/O & \scriptsize{1000} & \scriptsize{1000} & \scriptsize{1000} & \scriptsize{1200} & \scriptsize{1300} & \scriptsize{1500} & \scriptsize{1500}& \scriptsize{1200} & \scriptsize{1500} & \scriptsize{1500} & \scriptsize{1300} & \scriptsize{1500}\\
  \hline
  \hline
Multi10&\cmark&\cmark&\cmark&\cmark&\cmark&\cmark&\cmark&\cmark&\cmark&\cmark&\xmark&\xmark\\
Multi3&\xmark&\xmark&\xmark&\xmark&\xmark&\xmark&\xmark&\cmark&\cmark&\cmark&\xmark&\xmark\\
\hline
Suburban&\xmark&\xmark&\xmark&\xmark&\xmark&\xmark&\xmark&\cmark&\xmark&\xmark&\xmark&\xmark\\
Castle&\xmark&\xmark&\xmark&\xmark&\xmark&\xmark&\xmark&\xmark&\cmark&\xmark&\xmark&\xmark\\
TUM&\xmark&\xmark&\xmark&\xmark&\xmark&\xmark&\xmark&\xmark&\xmark&\cmark&\xmark&\xmark\\
Cal&\xmark&\xmark&\xmark&\xmark&\xmark&\xmark&\xmark&\xmark&\xmark&\xmark&\cmark&\xmark\\
Border&\xmark&\xmark&\xmark&\xmark&\xmark&\xmark&\xmark&\xmark&\xmark&\xmark&\xmark&\cmark\\
  \hline
Manhattan&\cmark&\xmark&\xmark&\xmark&\xmark&\xmark&\xmark&\xmark&\xmark&\xmark&\xmark&\xmark
\end{tabular}
    \caption{Table showing all maps indicating their respective sizes and timeout (T/O) values and detailing which agents (left column) were trained on which maps.}
    \label{tab:maps}
\end{table}

\begin{table}
    \centering
    \footnotesize
    \setlength{\tabcolsep}{2pt}
    \begin{tabular}{c|c|l||c|c|l}
     & Value & Description && Value & Description  \\ \hline
    $r_c$ & 0.01 & Coverage reward & $\epsilon$ & 0.1 & PPO clip\\
    $r_m$ & 0.02 & Motion penalty & $\lambda$ & 0.8 & TD($\lambda$) parameter \\
    $\gamma_0$ & 0.99 & Discount base & $|\theta|$ & 5\tiny{$\times 10^6$} & Actor parameters \\
    $\gamma_r$ & 0.1 & Discount decay rate & $|\phi|$ & 5\tiny{$\times 10^6$} & Critic parameters \\ 
    $\gamma_s$ & 2\tiny{$\times 10^7$} & Discount decay steps & - & 1.2\tiny{$\times 10^8$} & Interaction steps  \\\cline{4-6}
    \cline{1-3}
    $b_\text{max}$ & 100 & Max battery level &$g$ & 3 & Global map scaling \\
    $b_c$ & 2 & Charging amount &  $l$ & 17 & Local map size \\
    $\beta$ & 0.5 & Min initial charge &$\alpha$ & 0.99 & Position history decay  \\
    \end{tabular}
    \caption{Parameters.}
    \label{tab:parameters}
\end{table}
\section{Results}
\label{sec:results}
This section starts with ablation studies on action masking, discount factor scheduling, and position history. After that, an example shows how the coverage target is decomposed through the agent's trajectory, followed by detailed analysis of the generalization capabilities of the proposed approach. Note that all trajectory figures are clickable and lead to a timestamp in a video \footnote{\url{https://youtu.be/BeKn-VVhrz0}} showing an animation of that particular path.

\subsection{Action Masking Ablation}
\label{sec:mask_results}

This ablation study explores the impacts of different action masks on the learning performance. Multi3 agents (i.e., agents trained on three maps, see Table~\ref{tab:maps}) were trained using either no mask, the valid mask~\eqref{eq:invalid}, the immediate mask~\eqref{eq:immediate}, or the invariant mask~\eqref{eq:invariant}. For the purpose of this comparison, agents that violated a constraint received a penalty of $r_s = 5$ ($r_c=0.01$ and $r_m=0.02$), and the episode was terminated. Also specific to this ablation study, we found that the use of discount factor scheduling~\eqref{eq:discount} significantly deteriorated the performance of agents without the invariant mask. This can be attributed to the cumulative movement penalties surpassing the crash penalty as the discount factor increased. Consequently, for agents without the invariant mask, crashing became an optimal strategy when they had not yet learned to solve the task reliably. Therefore, the discount factor was set to a fixed value of $\gamma = 0.99$ for this ablation study.

\begin{figure}
    \centering
    \begin{subfigure}[t]{0.49\columnwidth}
        \includegraphics[width=\textwidth]{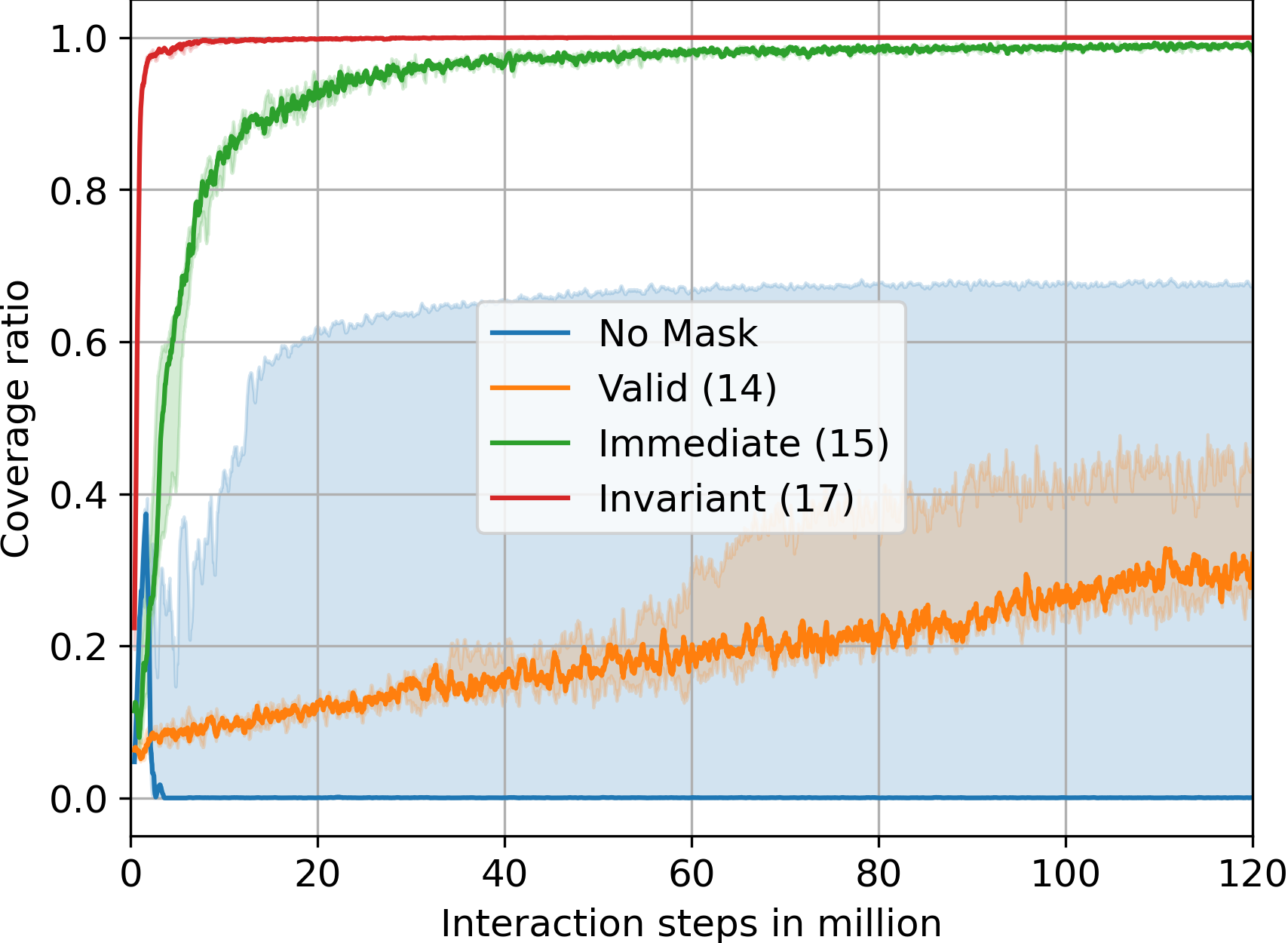}
        \caption{Average coverage ratio}
        \label{fig:mask_cov}
    \end{subfigure}%
    \hfill%
    \begin{subfigure}[t]{0.49\columnwidth}
        \includegraphics[width=\textwidth]{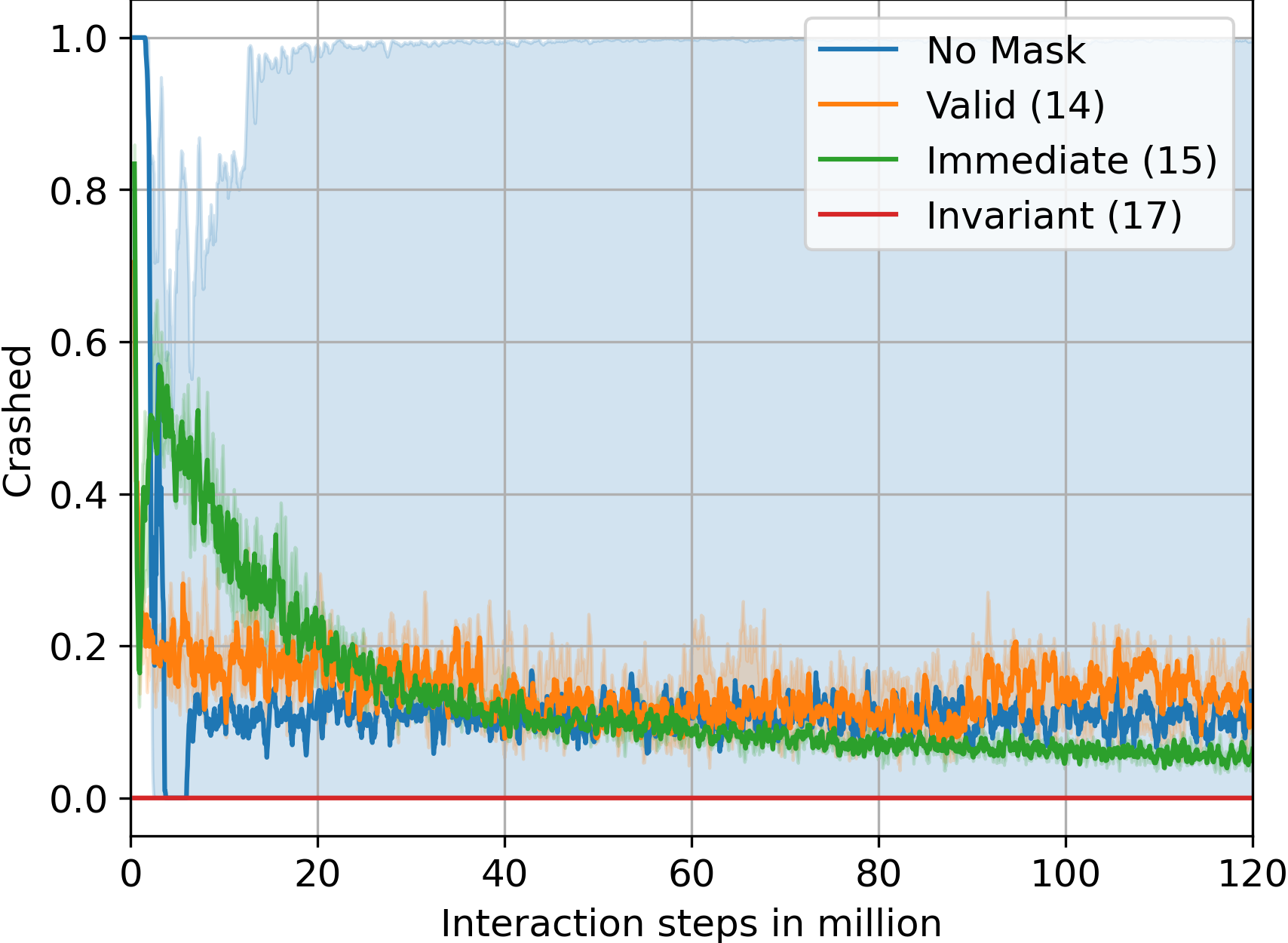}
        \caption{Ratio of crashes}
        \label{fig:mask_crash}
    \end{subfigure}\vspace{7pt}
    \begin{subfigure}[t]{0.49\columnwidth}
        \includegraphics[width=\textwidth]{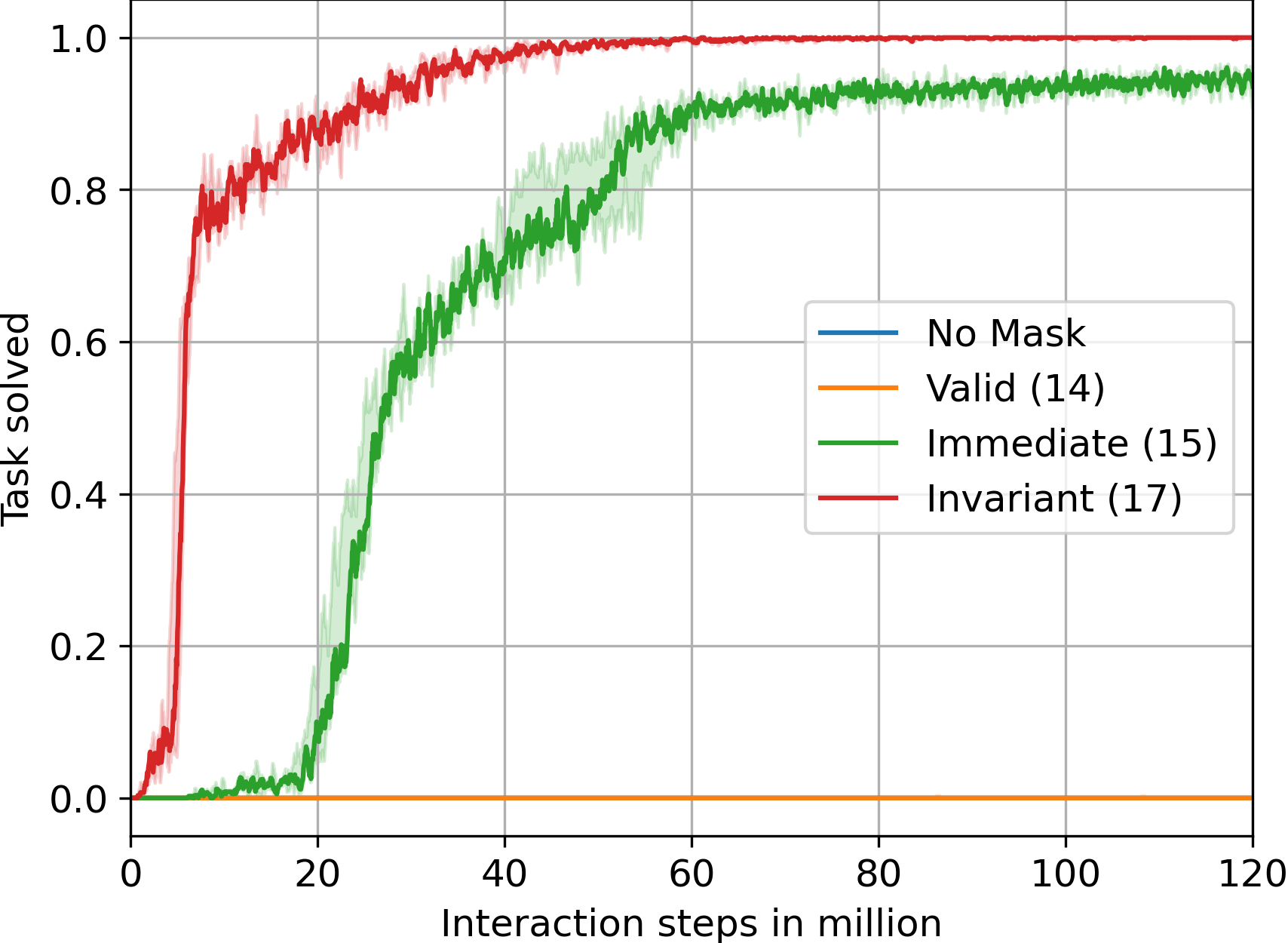}
        \caption{Ratio of solved tasks}
        \label{fig:mask_solved}
    \end{subfigure}
    \caption{Training curve using different action masks showing the median and min-max ranges of three agent training runs per masking approach.}
    \label{fig:masking}
\end{figure}

Figure~\ref{fig:masking} reports the training curves of the masking ablation, showing the average coverage ratio, constraint violation or crash ratio, and the ratio of solved tasks that are observed during the training rollouts. The invariant mask helps the agent to learn the task very early, which can be seen in the coverage ratio (Figure~\ref{fig:mask_cov}), and consequentially in the task solved indicator (Figure~\ref{fig:mask_solved}). Due to the safety guarantee, there are no safety violations during training, as seen in Figure~\ref{fig:mask_crash}. The agent with the immediate action mask learns to solve the task slower and less reliably. Even after 120 million interaction steps, only around 90-95\% of tasks were solved, and the rest resulted in a crash. The agent with the valid action mask learns to avoid crashing relatively fast, even though still showing a crash ratio of 10-20\% throughout training. However, the more significant problem is that the agent learns to cover the target region very slowly, leading to no solved task event throughout the training. Even worse, agents with no mask get stuck in one of two local optima. They fly out and cover as much as possible on one charge without recharging, or remain landed without covering anything. These two behaviors lead to the large variability in the indicators represented by the shaded blue region in the figures, with two agents exhibiting the remain landed behavior and the third agent venturing out with no return.

From this ablation, it can be concluded that more action masking yields increasing performance and safety. The only outliers in this safety trend are the agents without a mask that never take off to cover anything. Based on these results, we will only investigate agents with invariant action masks in the following. 

\subsection{Discount Factor Scheduling Ablation}
\label{sec:discount_results}
\begin{figure}
    \centering
    \begin{subfigure}[t]{0.48\columnwidth}
        \includegraphics[width=\textwidth]{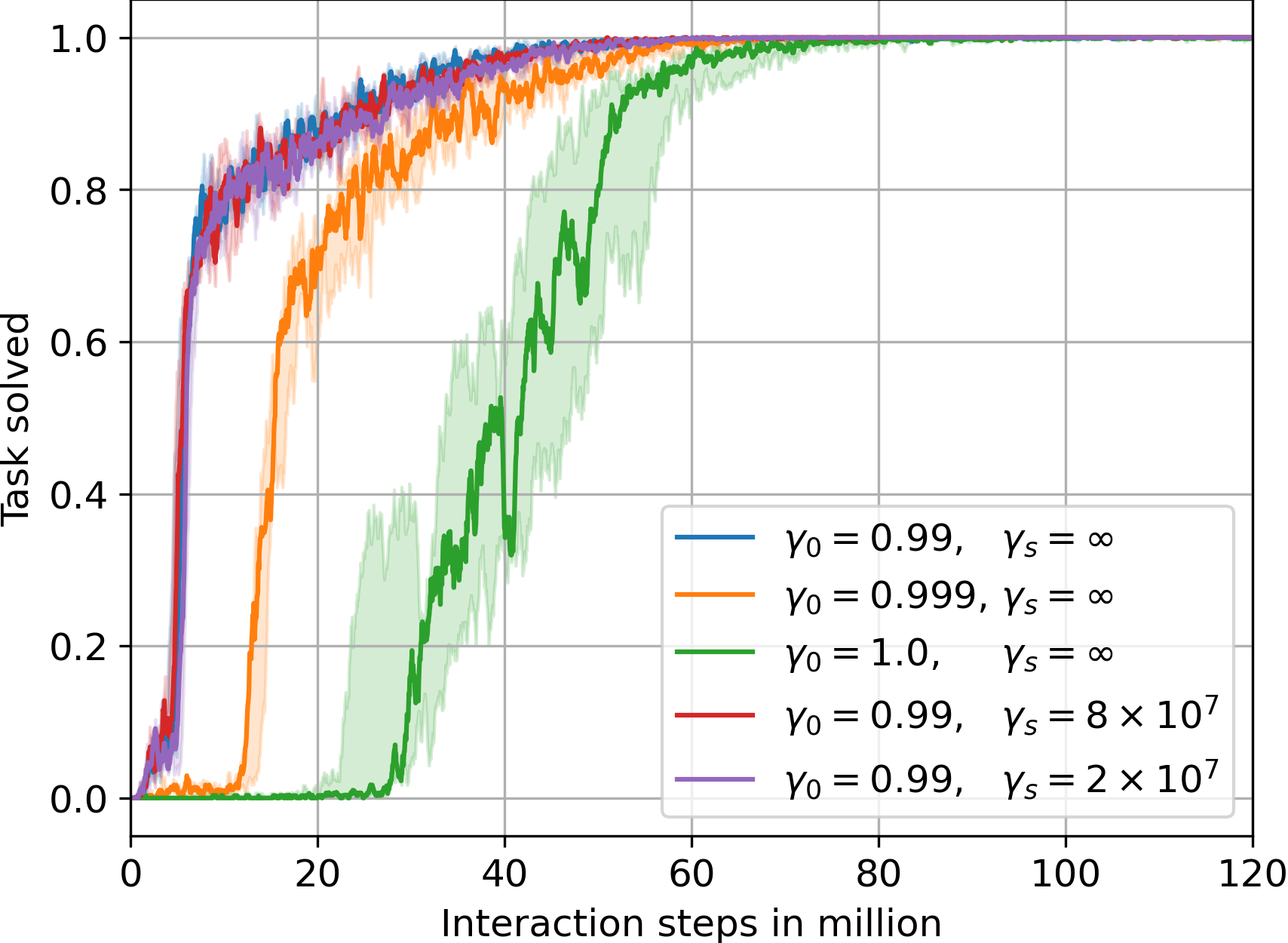}
        \caption{Ratio of solved tasks.}
        \label{fig:discount_solved}
    \end{subfigure}%
    \hfill%
    \begin{subfigure}[t]{0.505\columnwidth}
        \includegraphics[width=\textwidth]{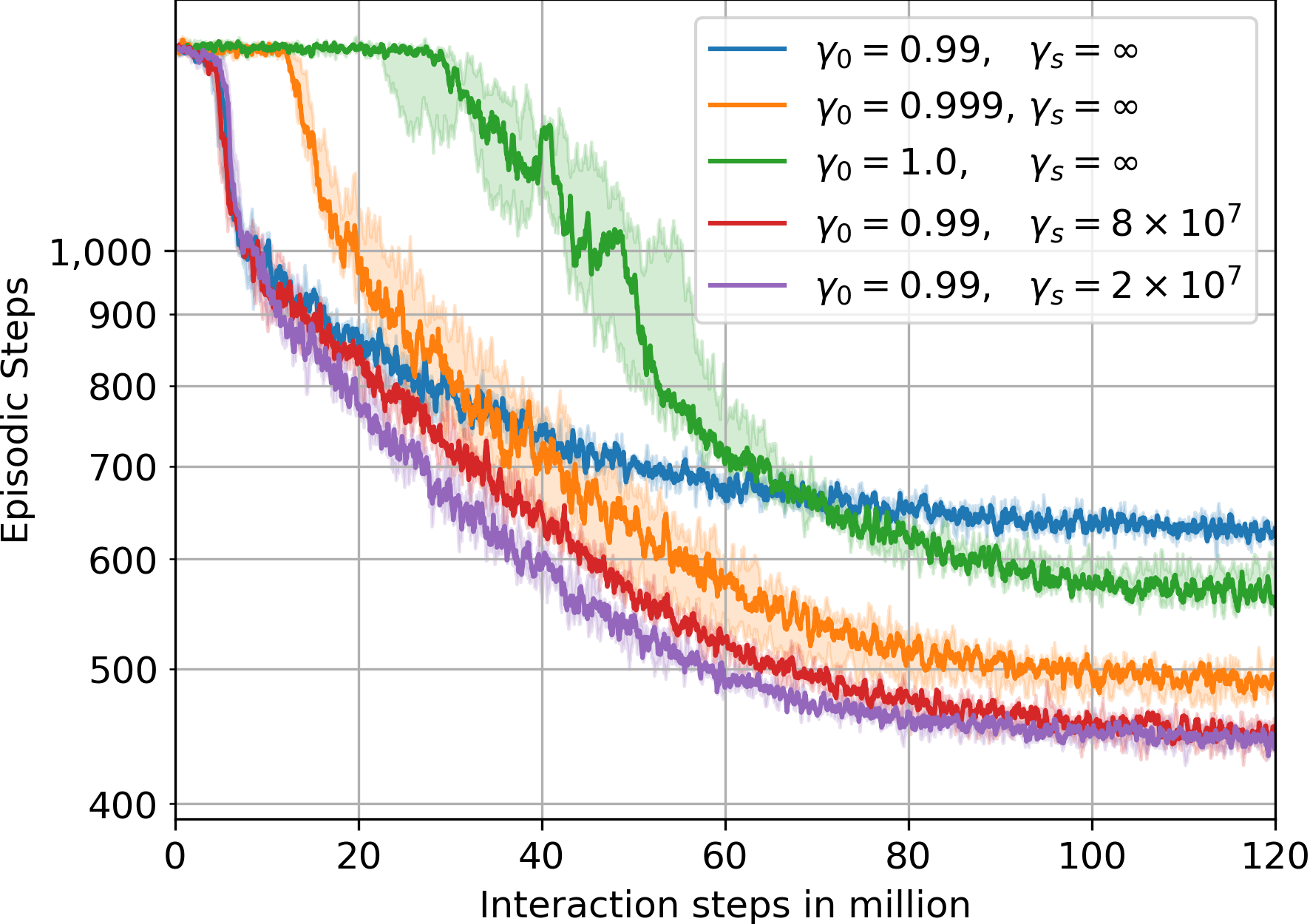}
        \caption{Episodic steps in log scale.}
        \label{fig:discount_steps}
    \end{subfigure}
    \caption{Training curve of different discount factor schedules.}
    \label{fig:discount}
\end{figure}
To investigate the effect of the proposed discount factor scheduling~\eqref{eq:discount}, we trained a number of Multi3 agents (see Table~\ref{tab:maps}) with different constant and scheduled discount factors. Specifically, three agents were trained with a constant discount factor $\{0.99, 0.999, 1.0\}$ and two with a scheduled discount factor starting from $0.99$ and going to $0.999$ in $\{2\times 10^7, 8\times 10^7\}$ steps. Figure~\ref{fig:discount} shows the ratio of solved tasks and steps per episode throughout training. It immediately becomes clear from Figure~\ref{fig:discount_solved} that overall higher constant discount factors lead to the agent learning to solve the task significantly later. Figure~\ref{fig:discount_steps} shows that the scheduled discount factor agents have a significant head start in minimizing the number of steps needed to finish each episode compared to the higher constant discount factors and a large advantage in final performance over the constant $0.99$ discount factor. The faster decay of the discount factor only aided in initial performance but did not yield better final performance. This ablation study confirms that having a lower discount factor aids in learning to solve the task and a higher one improves final performance. Combining the two through discount factor scheduling showed significant performance gains. Based on these results, we chose $\gamma_0=0.99$ and $\gamma_s=2\times 10^7$ for the following experiments.

\subsection{Position History Ablation}
\label{sec:pos_hist_results}

\begin{table}
\centering
\footnotesize
\setlength{\tabcolsep}{3pt}
\renewcommand{\arraystretch}{1.3}
\begin{tabular}{c|l||c|c||c|c|c}
& & \multicolumn{2}{c||}{\scriptsize Suburban + Castle + TUM} & \multicolumn{2}{c|}{Cal} & \multicolumn{1}{c}{Border}\\
&                  & solved       & RPD       & solved       & RPD & solved \\
\hline
\multirow{2}{1cm}{Base} & stoch.  &
$99.9^{0.0}$ & $-16.3^{0.5}$
&
$95.5^{2.7}$ & $9.7^{1.6}$ & $9.3^{0.9}$ 
 \\ \cline{2-7}
& deter.    &
$95.5^{0.5}$ & $-16.5^{0.7}$
&
$80.8^{6.1}$ & $10.4^{1.6}$ & $2.5^{0.8}$
 \\ \hline
\multirow{2}{1cm}{Random Layer} & stoch.  &
$\mathbf{100.0^{0.0}}$ & $-11.3^{0.9}$
&
$\mathbf{99.1^{0.6}}$ & $\mathbf{8.9^{1.3}}$ & $13.4^{3.2}$
 \\ \cline{2-7}
 & deter.    &
$99.7^{0.1}$ & $-11.6^{0.9}$
&
$98.5^{1.0}$ & $9.4^{1.3}$ & $11.5^{3.3}$
 \\ \hline
\multirow{2}{1cm}{Position History} & stoch.  &
$\mathbf{100.0^{0.0}}$ & $\mathbf{-17.6^{0.9}}$
&
$97.3^{0.6}$ & $10.2^{0.3}$ & $\mathbf{27.5^{3.1}}$
 \\ \cline{2-7}
 & deter.    &
$98.9^{0.3}$ & $-17.3^{0.9}$
&
$92.4^{0.7}$ & $12.8^{0.9}$ & $12.2^{1.9}$
\end{tabular}

\caption{Position history ablation comparing no history (base) with a random layer and position history on different maps showing \resizebox{!}{8pt}{$\{\text{mean}\}^{\{\text{std}\}}$} in \%.}\label{tab:history_ablation}
\end{table}

To evaluate the benefits of using the position history layer, we trained Multi3 agents (see Table~\ref{tab:maps}) with position history, without it, and with an added random layer. The random layer was added to see whether any changes in the observation were enough to break behavior loops (as described in section~\ref{subsec:position_history}). After training, the agents were evaluated on the maps they were trained on (Suburban, Castle, and TUM) and two unseen maps (Cal, and Border). For evaluation, $2^{10}$ scenarios per map were randomly generated. All agents were tested on these maps once deterministically with the $\operatorname{argmax}$ of the action logits and once sampling from a $\operatorname{softmax}$ distribution of the action logits stochastically.

The results in Table~\ref{tab:history_ablation} show the performance in solving the task and the RPD compared to the heuristic~\eqref{eq:rpd} of each configuration averaged over the trained maps, and the Cal and Border maps separately. For the Border map, the RPD values are omitted as they are not comparable for low task-solved ratios. On in-distribution maps, the deterministic base case shows that 4.5\% of the maps are not solved, most likely due to loops. The loop problem seems solvable by making the agent stochastic, adding a random layer, or adding the position history layer. The deterministic agent with position history does not always solve the task but is significantly better than the deterministic base agent. In RPD, the random layer decreases performance, while the position history improves performance. For the out-of-distribution Cal map, the difference in task solving is more significant with the best performance of the stochastic agent with random layer. Interestingly, on the Cal map, the random layer shows the best RPD performance. However, the most striking difference between the approaches can be seen on the Border map. The stochastic position history agent solves more than double of the scenarios than the other approaches.

While sampling from the $\operatorname{softmax}$ distribution solves the loop problem for in-distribution maps, the position history significantly improved task completion for the out-of-distribution maps. Additionally, the position history improves performance for in-distribution maps, yielding the best performance. We hypothesize that the stochastic actions allow the agent to \textit{break out} of loops while the position history \textit{prevents} the loops in the first place. Additionally, for the complicated Border map, the agent might use the position history to \textit{remember} what it tried before, which allows it to solve significantly more scenarios. Given this knowledge, the following results perform stochastic inference using the position history layer.

\subsection{Target Decomposition Example}
\begin{figure}
    \centering
    \href[pdfnewwindow]{https://youtu.be/BeKn-VVhrz0?t=25}{
    \begin{subfigure}{0.49\columnwidth}
        \centering
        \includegraphics[width=\textwidth]{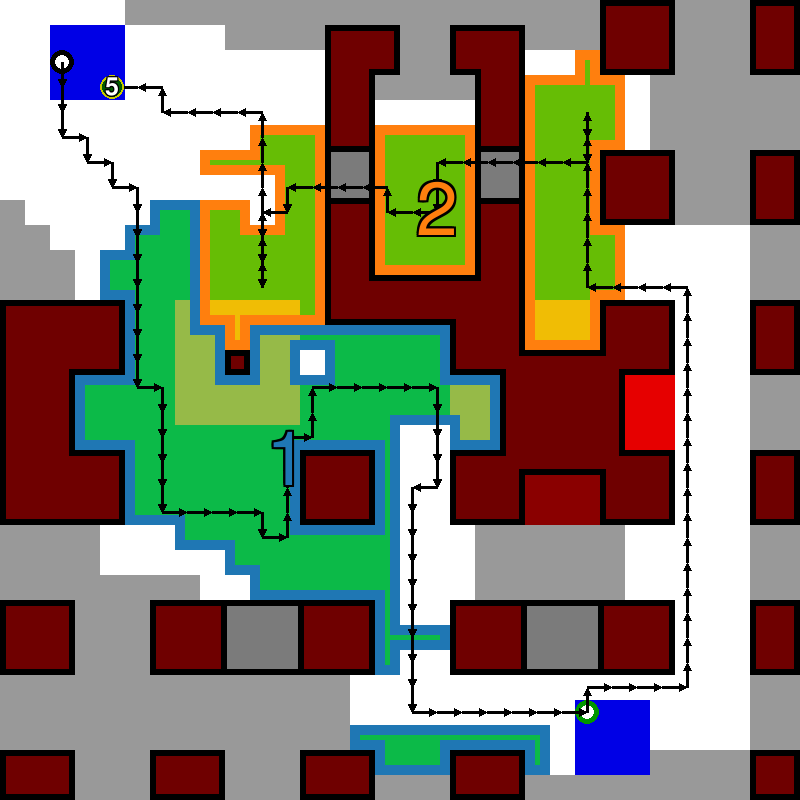}
        \caption{Manhattan agent: 166 steps}\label{fig:decomp2}
    \end{subfigure}\hfill%
    \begin{subfigure}{0.49\columnwidth}
        \centering
        \includegraphics[width=\textwidth]{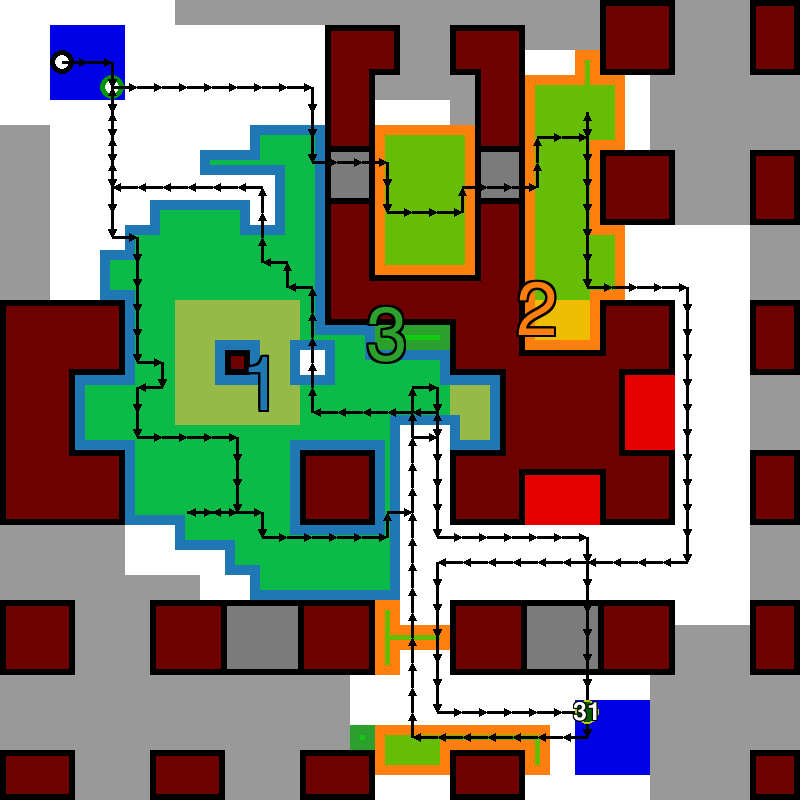}
        \caption{Greedy heuristic: 266 steps}\label{fig:decomp2_heur}
    \end{subfigure}%
    }
    \href[pdfnewwindow]{https://youtu.be/BeKn-VVhrz0?t=50}{
    \begin{subfigure}{0.49\columnwidth}
        \centering
        \includegraphics[width=\textwidth]{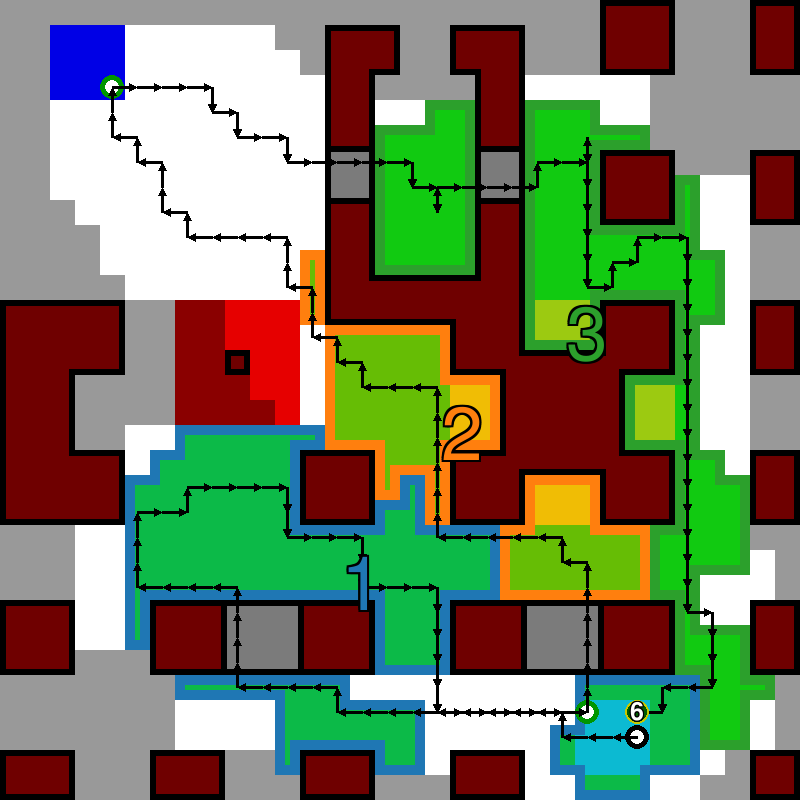}
        \caption{Manhattan agent: 225}\label{fig:decomp3}
    \end{subfigure}\hfill%
    \begin{subfigure}{0.49\columnwidth}
        \centering
        \includegraphics[width=\textwidth]{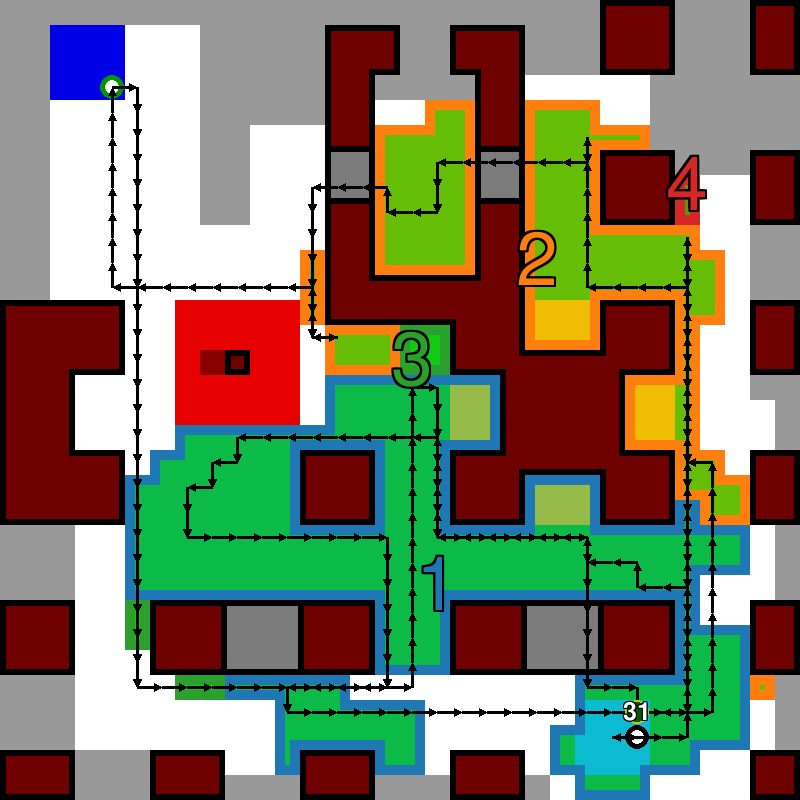}
        \caption{Greedy heuristic: 382}\label{fig:decomp3_heur}
    \end{subfigure}%
    }
    \caption{Two examples of resulting target zone decomposition between multiple passes of the agent and the heuristic, with hyperlinks behind the graphics leading to videos of the trajectories.}
    \label{fig:decompositions}
\end{figure}
In contrast to most existing works on coverage path planning, we investigate scenarios with a hard energy constraint that makes it necessary for the UAV agent to learn when and how to integrate recharging stops into its trajectory. To illustrate the effect of the capability to recharge on the resulting trajectory, Figure~\ref{fig:decompositions} shows a Manhattan agent and the greedy heuristic solving two scenarios on the Manhattan map. For this example, during training and testing, the maximum battery level was set to $b_\text{max}=75$, so that solving the problem in one pass is rarely possible. The numbered areas that are outlined and shaded in different colors show in which pass the agent is covering that area, with a recharge stop between every pass. 

Focusing first on the Manhattan agent, in Figure~\ref{fig:decomp2}, the agent starts in the top-left landing zone, and flies down to the bottom-right landing zone to recharge, covering the blue-shaded area. After recharging, it flies around the top-right to cover the remaining area and land on the top-left landing zone. In the scenario in Figure~\ref{fig:decomp3}, the target area is too large to cover in two passes. Hence, the agent splits the target into three passes, starting from the bottom-right landing zone and consecutively landing on the bottom-right, top-left, and bottom-right zones.

The greedy heuristic on the other hand requires significantly more steps than the DRL agent. The reason is primarily that it does not optimize the timing for recharge trips, leaving behind small patches of the target zone, requiring additional passes. This can be seen by the third target decomposition in Figure~\ref{fig:decomp2_heur} and the third and fourth decompositions in Figure~\ref{fig:decomp3_heur}. This example illustrates the challenge in the \gls{cpp} problem with recharge: the agent needs to precisely predict which cells it can cover in the current and in future passes. Creating these decompositions with traditional techniques would be challenging. However, using \gls{rl}, the agent appears to learn implicitly how to decompose the target zones.

\subsection{Generalization}

\begin{table*}
\def\i{\cellcolor{SpringGreen}}
\def\o{\cellcolor{SkyBlue}}
    \centering
\footnotesize
\setlength{\tabcolsep}{2pt}
\renewcommand{\arraystretch}{1.3}
\newcommand\diagfil[4]{%
  \multicolumn{1}{p{#1}|}{\hskip-\tabcolsep
  $\vcenter{\begin{tikzpicture}[baseline=0,anchor=south west,inner sep=0pt,outer sep=0pt]
  \path[use as bounding box] (0,0) rectangle (#1+2\tabcolsep,\baselineskip);
  \node[minimum width={#1+2\tabcolsep},minimum height=\arraystretch\baselineskip,fill=#2] (box)at(0,-1pt) {};
  \fill [#3] (box.south west)--(box.north east) |- cycle;
  \node[anchor=center] at (box.center) {#4};
  \end{tikzpicture}}$\hskip-\tabcolsep}}
\newcommand\iosol[1]{\diagfil{1.13cm}{SpringGreen}{SkyBlue}{#1}}
\newcommand\iorpd[1]{\diagfil{1.26cm}{SpringGreen}{SkyBlue}{#1}}
  
\begin{tabular}{c||c|c|c|c|c|c||c|c|c|c|c|c}
 Maps $\rightarrow$ & \multicolumn{2}{c|}{Suburban} & \multicolumn{2}{c|}{Castle} & \multicolumn{2}{c||}{TUM} & \multicolumn{2}{c|}{Castle\textsuperscript{*}} &\multicolumn{2}{c|}{Cal}  &\multicolumn{2}{c}{Border} \\
 Agents $\downarrow$ & solved & RPD & solved & RPD & solved & RPD & solved & RPD & solved & RPD & solved & RPD \\
\hline
Multi10 & 
$\mathbf{100.0^{0.0}}$ \i & $-8.9^{0.0}$ \i & $\mathbf{100.0^{0.0}}$ \i & $-10.7^{0.7}$ \i & $\mathbf{100.0^{0.0}}$ \i & $-7.1^{0.9}$ \i & \iosol{$\mathbf{100.0^{0.0}}$} & \iorpd{$-7.4^{0.2}$} \i & $99.3^{0.6}$ \o & $-1.6^{2.5}$ \o & $35.0^{2.6}$ \o & $21.5^{1.7}$ \o
 \\ \hline
Multi3 & 
$\mathbf{100.0^{0.0}}$ \i & $-15.4^{0.6}$ \i & $\mathbf{100.0^{0.0}}$ \i & $-18.3^{0.6}$ \i & $\mathbf{100.0^{0.0}}$ \i & $-19.2^{1.6}$ \i & \iosol{$\mathbf{100.0^{0.0}}$} & \iorpd{$-12.3^{0.9}$} & $97.3^{0.6}$ \o & $10.2^{0.3}$ \o & $27.5^{3.1}$ \o & $37.9^{4.5}$ \o
 \\ \hline\hline
Suburban & 
$\mathbf{100.0^{0.0}}$ \i & $\mathbf{-28.7^{0.5}}$ \i & $1.0^{0.7}$ \o & $-$ \o & $0.9^{0.5}$ \o & $-$ \o & $2.9^{2.8}$ \o & $-$ \o & $25.1^{20.3}$ \o & $126.2^{11.6}$ \o & $0.2^{0.2}$ \o & $-$ \o
 \\ \hline
Castle & 
$33.3^{2.9}$ \o & $65.5^{7.5}$ \o & $\mathbf{100.0^{0.0}}$ \i & $\mathbf{-23.7^{0.3}}$ \i & $2.6^{0.6}$ \o & $-$ \o & \iosol{$\mathbf{100.0^{0.0}}$} & \iorpd{$\mathbf{-16.2^{0.3}}$} & $84.6^{9.3}$ \o & $23.7^{8.2}$ \o & $9.4^{5.1}$ \o & $-$ \o
 \\ \hline
TUM & 
$5.6^{2.5}$ \o & $-$ \o & $0.3^{0.2}$ \o & $-$ \o & $\mathbf{100.0^{0.0}}$ \i & $\mathbf{-26.5^{0.4}}$ \i & $1.4^{1.2}$ \o & $-$ \o & $94.9^{2.4}$ \o & $23.8^{2.2}$ \o & $0.3^{0.3}$ \o & $-$ \o
 \\ \hline
Cal & 
$10.6^{7.4}$ \o & $102.6^{25.6}$ \o & $0.6^{0.4}$ \o & $-$ \o & $1.7^{0.5}$ \o & $-$ \o & $1.3^{1.3}$ \o & $-$ \o & $\mathbf{100.0^{0.0}}$ \i & $\mathbf{-30.7^{0.2}}$ \i & $0.1^{0.2}$ \o & $-$ \o
 \\ \hline
Border & 
$60.6^{8.3}$ \o & $59.5^{7.7}$ \o & $9.3^{1.7}$ \o & $-$ \o & $14.6^{6.0}$ \o & $55.2^{3.2}$ \o & $20.0^{6.1}$ \o & $45.1^{6.6}$ \o & $93.4^{0.7}$ \o & $20.7^{0.8}$ \o & $\mathbf{100.0^{0.0}}$ \i & $\mathbf{-28.6^{0.1}}$ \i
\end{tabular}
\caption{Performance comparison of different agents on scenarios of all maps. If the task is solved for fewer than 10\% of scenarios, RPD values are irrelevant and not listed. The values show \resizebox{!}{8pt}{$\{\text{mean}\}^{\{\text{std}\}}$} in \% and the cell color indicates if the agent was \textcolor{SpringGreen}{\textbf{trained on}} that map or if it is \textcolor{SkyBlue}{\textbf{unseen}}. The Castle\textsuperscript{*} map is a slight modification of the original map (see Figure~\ref{fig:castle2_comp}), which was neither directly trained on nor was it completely unseen during training. Lower RPD values are better.}
    \label{tab:comparison}
\end{table*}

To analyze how well the agents can generalize their learned abilities, we evaluate the trained agents extensively on multiple maps. All results are shown in Table~\ref{tab:comparison} and discussed in detail in the following. We concentrate on three aspects: the generalization over target zones, multiple maps, and unseen maps.

\subsubsection{Target Zones}

\begin{figure}
    \centering
    \href[pdfnewwindow]{https://youtu.be/BeKn-VVhrz0?t=83}{
    \begin{subfigure}{0.48\columnwidth}
        \includegraphics[width=\textwidth]{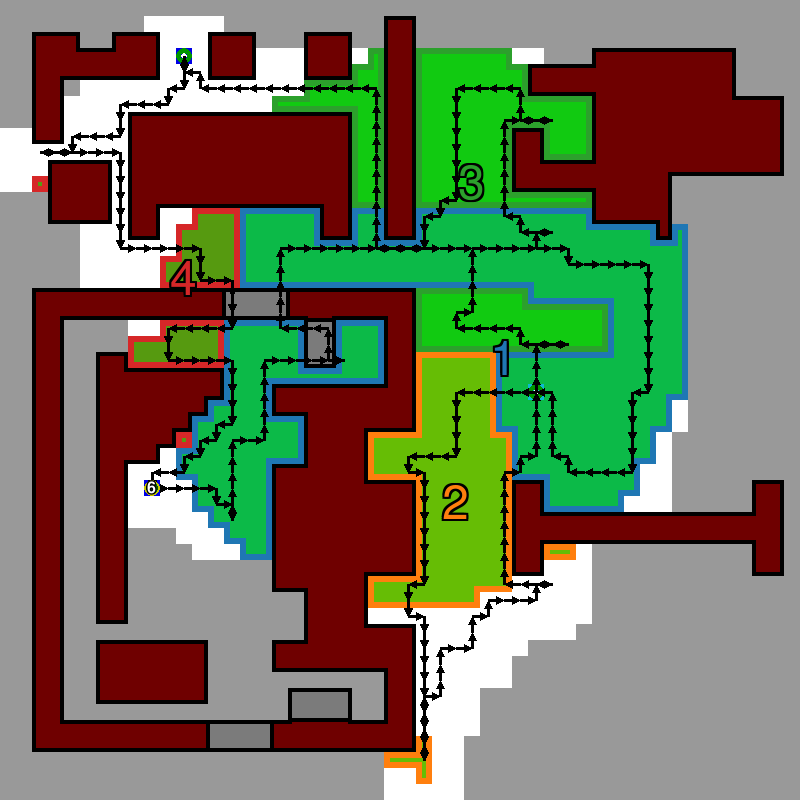}
        \caption{Solved after 422 steps.}
    \end{subfigure}\hspace{5pt}%
    \begin{subfigure}{0.48\columnwidth}
        \begin{tikzpicture}
            \node[inner sep=0pt] {\includegraphics[width=\textwidth]{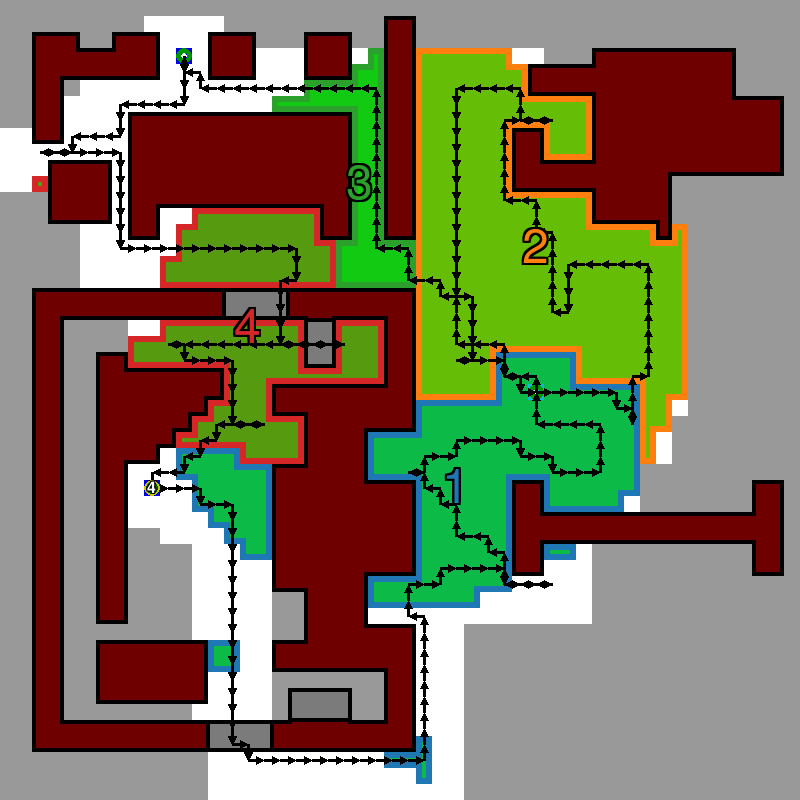}};
            \draw [->, ultra thick, green] (-2,-2) -- (-1.0,-1.35);         
        \end{tikzpicture}        
        \caption{Solved after 424 steps.}
    \end{subfigure}%
    }
    \caption{Two scenarios on the TUM map that are only different by a small target zone indicated with the green arrow, leading to significantly different trajectories of the TUM agent.}
    \label{fig:tum_small_change}
\end{figure}

For the target zone generalization, we focus on the five specialized agents at the bottom of Table~\ref{tab:comparison} evaluated on the same map they each have been trained on indicated by the \textcolor{SpringGreen}{\textbf{green color}}. The respective maps (landing zones, NFZs, and obstacles) are the same during training and evaluation in these evaluations. However, the initial position, battery, and target zones are randomly sampled from their respective distributions. While the space of initial positions and battery levels is relatively small, the space of random target zones is vast. Therefore, in each episode, the agent encounters a different, unseen target zone sampled from the same distribution, requiring \textit{in-distribution} generalization. 

In Figure~\ref{fig:tum_small_change}, we show an example of the impact of a minor change in the target zone on the agent's trajectory. Specifically, a small target zone was added in the scenario on the right, changing the agent's trajectory almost entirely. By flying the top route in the unchanged scenario on the left, the agent required two fewer steps than the bottom route in the modified scenario on the right. Since the agent's trajectory appears to be very sensitive to the target zones, generalization over different target zones can be considered challenging.

Considering the performance displayed by the specialized agents (Suburban, Castle, TUM, Cal, Border) on their respective training maps (see Table~\ref{tab:comparison}), it is clear that the agents always learn to solve their respective training maps efficiently. Their RPD to the baseline heuristic shows an average reduction of 24-31\% in the number of required steps to solve the problem. It can be concluded that the agents can generalize solving the coverage problem for randomized target distributions and significantly outperform the heuristic.

\subsubsection{Multimap}

\begin{figure}
\centering
\def\x{0.98}
    \href[pdfnewwindow]{https://youtu.be/BeKn-VVhrz0?t=118}{
\begin{subfigure}{0.5\columnwidth}
    \centering
    \includegraphics[width=\x\textwidth]{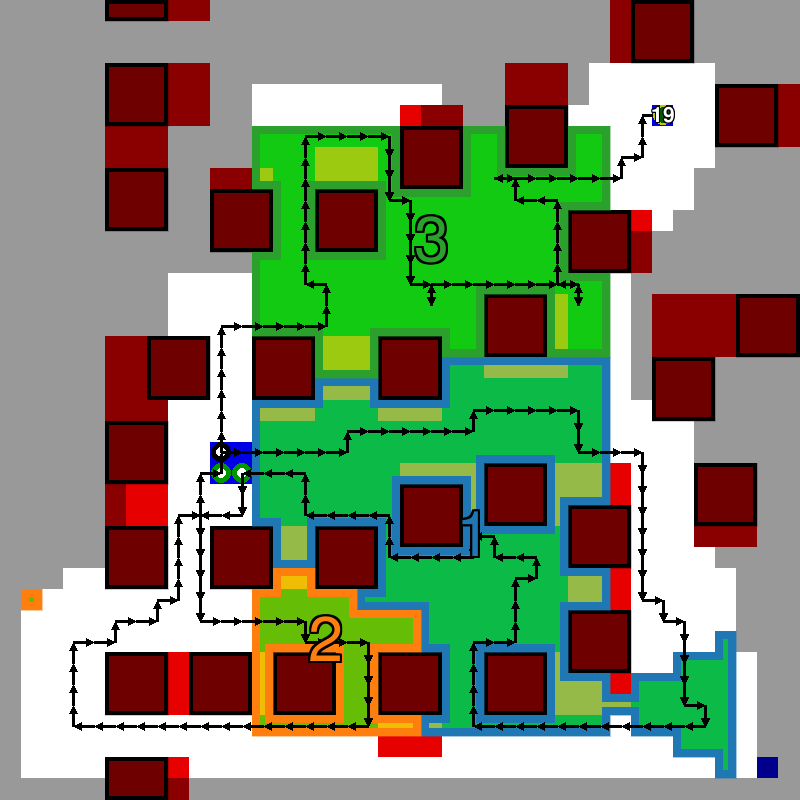}
    \caption{Multi3 agent: 281 steps}\label{fig:multi3_suburban}
\end{subfigure}%
\begin{subfigure}{0.5\columnwidth}
    \centering
    \includegraphics[width=\x\textwidth]{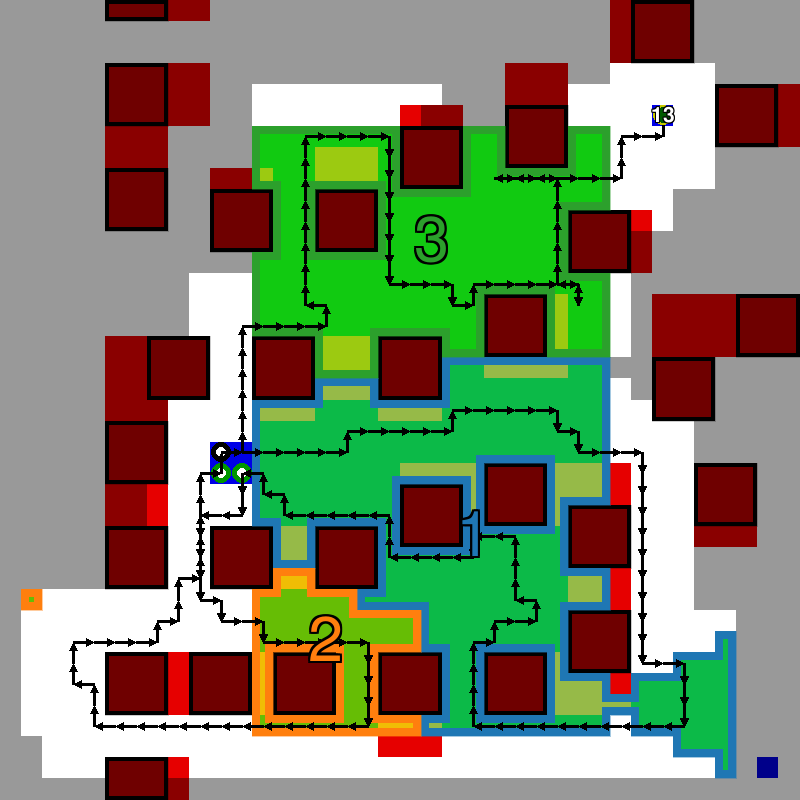}
    \caption{\mbox{Specialized agent: 275 steps}}\label{fig:suburban_suburban}
\end{subfigure}%
}
\href[pdfnewwindow]{https://youtu.be/BeKn-VVhrz0?t=144}{
\begin{subfigure}{0.5\columnwidth}
    \centering
    \includegraphics[width=\x\textwidth]{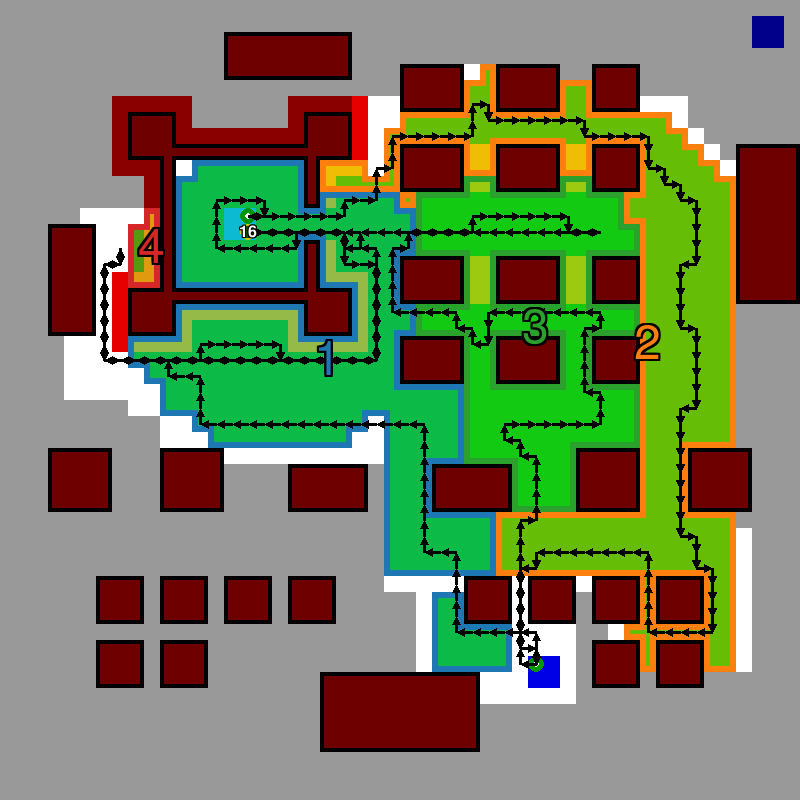}
    \caption{Multi3 agent: 524 steps}\label{fig:multi3_castle}
\end{subfigure}%
\begin{subfigure}{0.5\columnwidth}
    \centering
    \includegraphics[width=\x\textwidth]{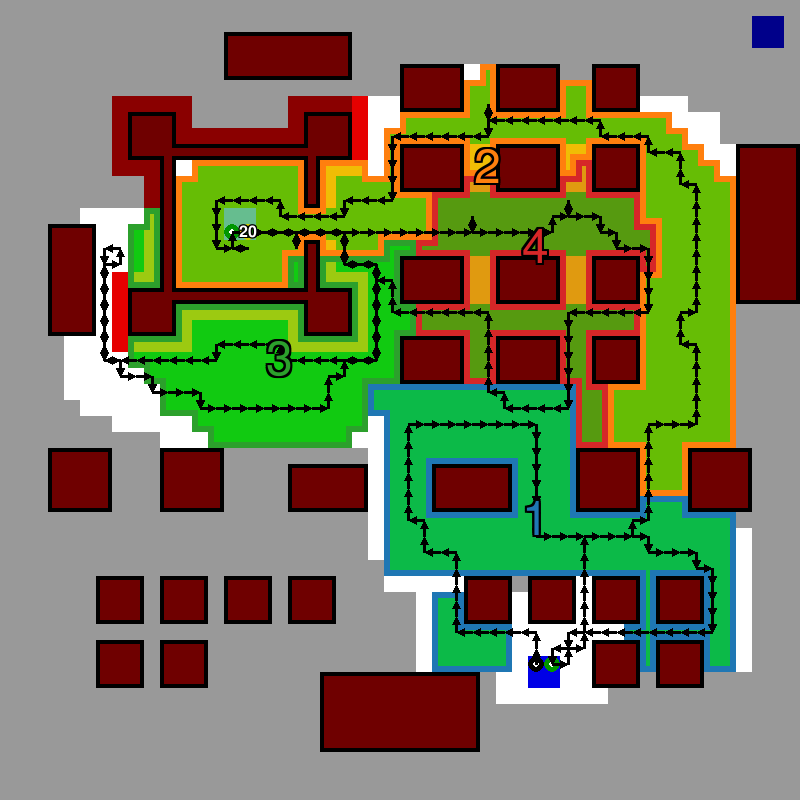}
    \caption{Specialized agent: 485 steps}\label{fig:castle_castle}
\end{subfigure}%
}
\href[pdfnewwindow]{https://youtu.be/BeKn-VVhrz0?t=186}{
\begin{subfigure}{0.5\columnwidth}
    \centering
    \includegraphics[width=\x\textwidth]{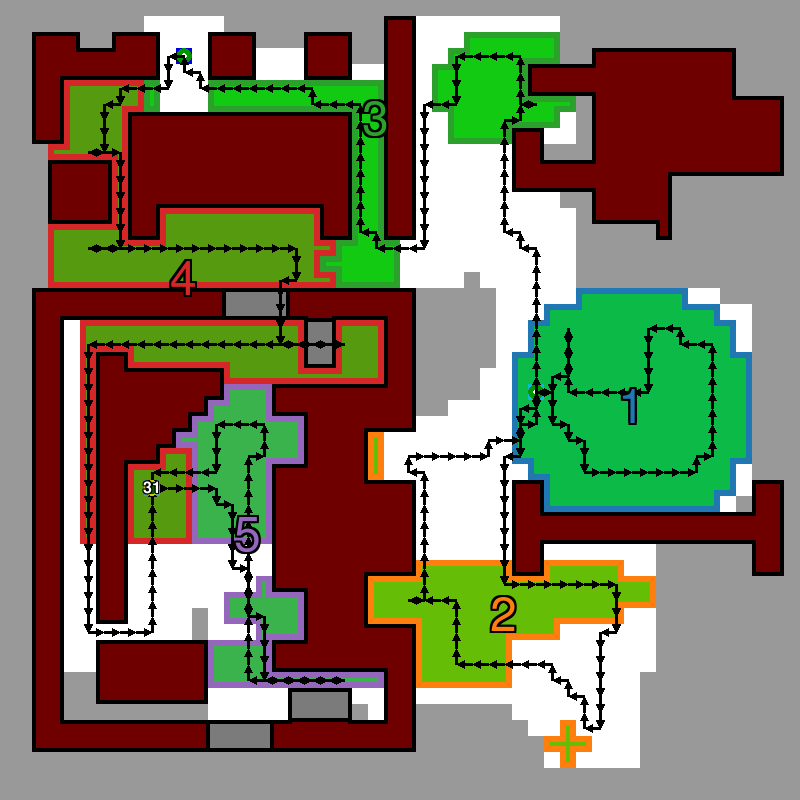}
    \caption{Multi3 agent: 511 steps}\label{fig:multi3_tum}
\end{subfigure}%
\begin{subfigure}{0.5\columnwidth}
    \centering
    \includegraphics[width=\x\textwidth]{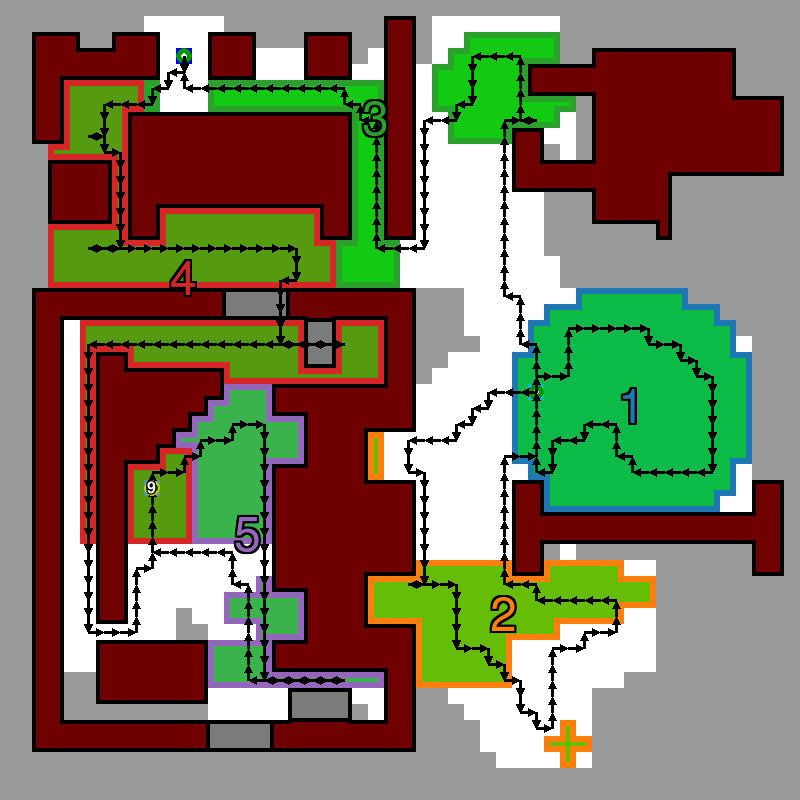}
    \caption{Specialized agent: 494 steps}\label{fig:tum_tum}
\end{subfigure}%
}
\caption{Comparison trajectories, showing the same problem solved by the Multi3 and the respective specialized agents for the Suburban, Castle, and TUM maps.}
\label{fig:comparison}
\end{figure}

For multimap generalization capabilities, we focus on the results of the Multi10 and Multi3 agents at the top of Table~\ref{tab:comparison} on the first three maps (Suburban, Castle, TUM). The main questions are whether an agent can learn to generalize over target zones on multiple maps and, if so, how large the performance loss is compared to specialized agents. As such, successful evaluation on multiple maps (that the agent was trained on) requires \textit{in-sample} generalization for the maps and \textit{in-distribution} generalization for target zones.

Figure~\ref{fig:comparison} shows example trajectory comparisons between the Multi3 agent, which was trained on all three example maps, and the respective specialized agents that each were only trained on one map (Suburban, Castle, or TUM). For all three examples, the specialized agents found a shorter trajectory than the Multi3 agent. On the Suburban map (Figure~\ref{fig:multi3_suburban}+~\ref{fig:suburban_suburban}), the agents created identical decompositions, with slight deviations along the trajectory. The specialized Suburban agent required six fewer steps, partly because it arrived with less remaining battery at the landing zone. On the Castle map (Figure~\ref{fig:multi3_castle} +~\ref{fig:castle_castle}), the agents show different trajectories, with the most significant improvement by the specialized agent being that it added the decomposition region 4 of the Multi3 agent into its region 3, which lead to a significantly shorter path. On the TUM map (Figure~\ref{fig:multi3_tum} +~\ref{fig:tum_tum}), the agents again created nearly identical decompositions. However, on this map, the difference in steps is greater, making the specialized agent's trajectory 17 steps shorter. A significant contributor is the difference in the final battery level of 22, \mbox{i.e., the equivalent of 11 charging steps. }

Quantitative results in Table~\ref{tab:comparison} confirm that the Multi agents perform worse than the specialized agents on their respective maps. The Multi10 agent shows a more significant relative performance decrease than the Multi3 agent. However, both agents still achieve a perfect task-solved ratio, and all RPD values are still negative, showing that the agents outperform the heuristic. The drop in performance for the Multi agents is expected since training steps and size of the neural networks remain unchanged compared to the specialized agents.

\subsubsection{Unseen Maps}
For the unseen map generalization, we focus on all cells of Table~\ref{tab:comparison} with the \textcolor{SkyBlue}{\textbf{blue color}} with a primary focus on the Multi agents. In these evaluations, the agents are deployed in maps they did not encounter during training. To solve unseen maps, the agents require \textit{out-of-distribution} generalization as the scenarios encountered are not within the distributions drawn from during training. The unseen maps include a slight modification of the Castle map, the relatively simple Cal map, and the challenging Border map. For the specialized agents, the unseen maps also include all other maps except the one each specialized agent was trained on.

\begin{figure}
    \centering
    \href[pdfnewwindow]{https://youtu.be/BeKn-VVhrz0?t=227}{
    \begin{subfigure}{0.5\columnwidth}
        \centering
        \includegraphics[width=0.98\textwidth]{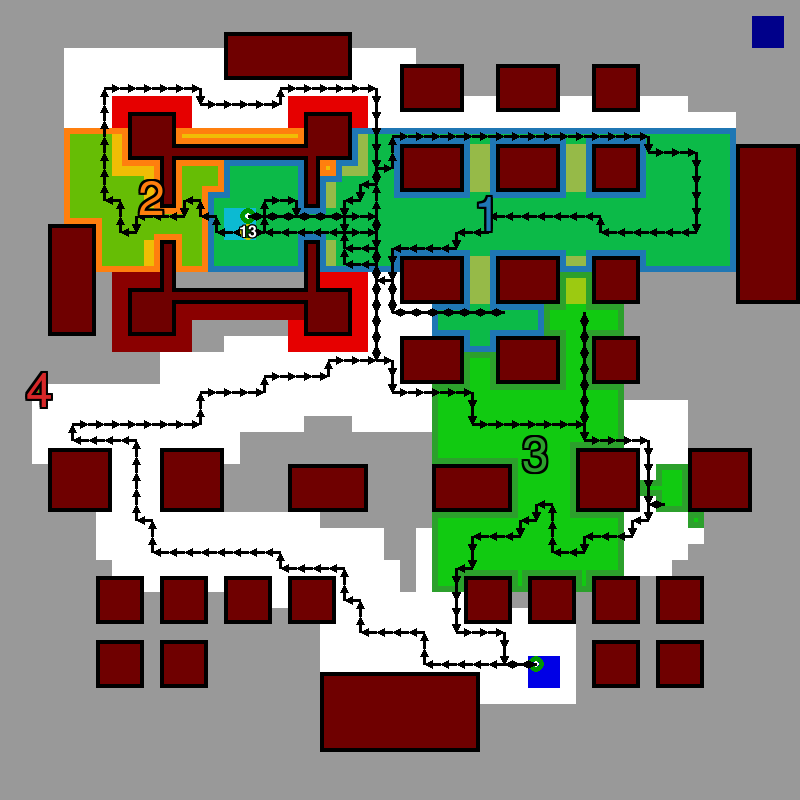}
        \caption{Multi10 agent: 473 steps}\label{fig:multi10_castle2}
    \end{subfigure}%
    \begin{subfigure}{0.5\columnwidth}
        \centering
        \includegraphics[width=0.98\textwidth]{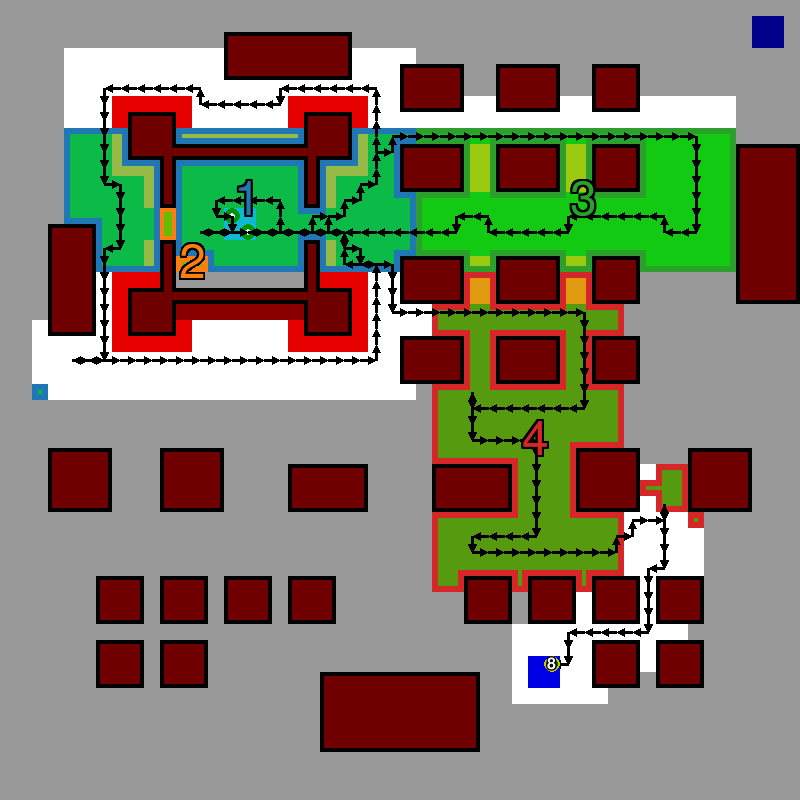}
        \caption{Castle agent: 374 steps}\label{fig:castle_castle2}
    \end{subfigure}%
    }
    \caption{Trajectories generated by the Multi10, and Castle agents on the modified Castle\textsuperscript{*} map. The castle on the top left has an added entrance on the left, which does not exist on the original map encountered during training.}
    \label{fig:castle2_comp}
\end{figure}

Figure~\ref{fig:castle2_comp} shows an example of how the Multi10 and Castle agents react to a slight modification of the Castle map that both agents have encountered during training. In contrast to the original, the modified Castle\textsuperscript{*} map includes a second entrance to the castle obstacle on the top left. In the example scenario shown, it would be advantageous for the agent to use the second castle entrance to easily reach the target zone on the top left. It can be seen that the specialized Castle agent ignores the new possibility and solves the scenario as it would solve it on the unaltered map. On the other hand, the Multi10 agent adapts and takes advantage of the second entrance, making this part of the trajectory more efficient. However, it still requires more steps than the specialized Castle agent because the rest of the trajectory is less optimized. Unfortunately, the Multi10 agent does not reliably take advantage of this map simplification but sometimes does, in contrast to the Castle and Multi3 agents, which never utilize the second entrance. This can be seen in Table~\ref{tab:comparison}, where all agents' RPD performance drops compared to the original Castle map, as the heuristic takes advantage of the second entrance. As expected, the task-solved ratios remain unaffected. Notably, the Border agent solves the modified Castle\textsuperscript{*} map twice as often as the original Castle map, showing that the former is easier than the latter. 

\begin{figure}
    \centering
    \href[pdfnewwindow]{https://youtu.be/BeKn-VVhrz0?t=265}{
    \begin{subfigure}{0.5\columnwidth}
        \centering
        \includegraphics[width=0.98\textwidth]{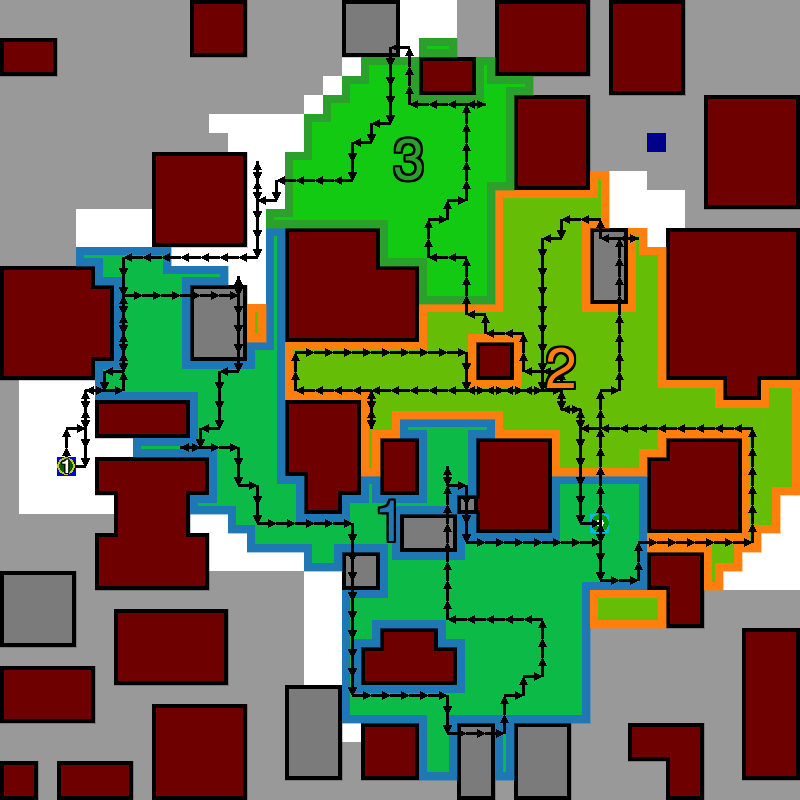}
        \caption{Multi10 agent: 393 steps}\label{fig:multi10_cal}
    \end{subfigure}%
    \begin{subfigure}{0.5\columnwidth}
        \centering
        \includegraphics[width=0.98\textwidth]{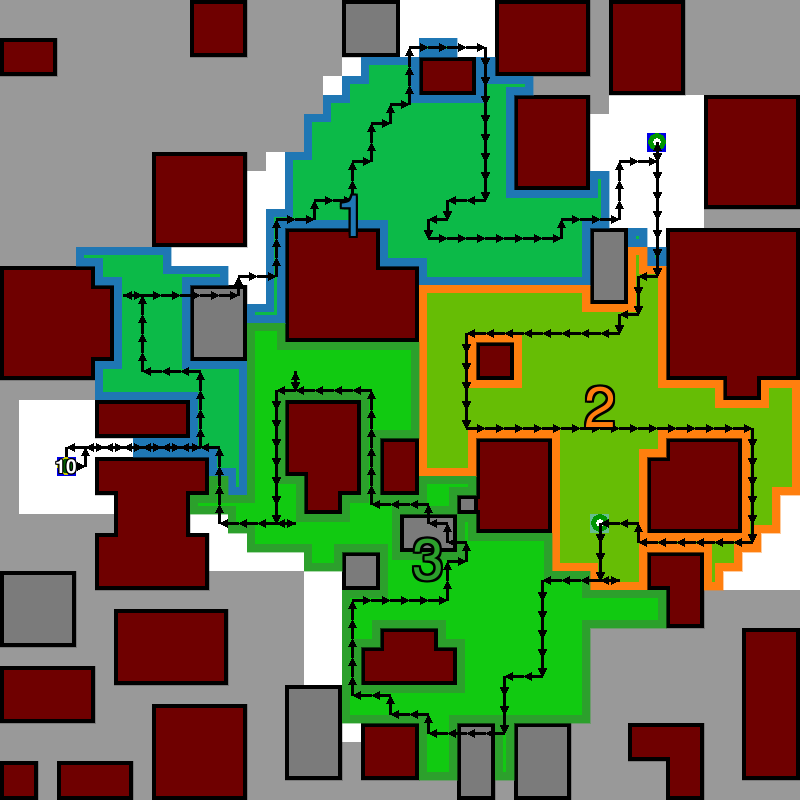}
        \caption{Specialized agent: 313 steps}\label{fig:cal_cal}
    \end{subfigure}%
    }
    \caption{Trajectories generated by the Multi10, and specialized Cal agents on the Cal map as an example scenario.}
    \label{fig:cal_comp}
\end{figure}

The Cal map is the first one that has not been seen at all by any agent during training, except for the Cal specialized agent. Example trajectories of the Multi10 and Cal agent are shown in Figure~\ref{fig:cal_comp}. It is evident that the Multi10 agent can solve the Cal map but takes significantly longer than the specialized agent. Referring to the overall results in the Cal column in Table~\ref{tab:comparison}, it is clear that the Cal map is indeed relatively easy as the task-solved ratio of all agents is higher than for any other map in the table. The TUM and Border agents specialized on different maps even solved it in $>90\%$ of all evaluated scenarios. However, it can be clearly seen that the Multi agents perform best in applying their generalized trajectory planning abilities to this new map, with the Multi10 agent producing better results than the heuristic. 

\begin{figure}
    \centering
    \href[pdfnewwindow]{https://youtu.be/BeKn-VVhrz0?t=299}{
    \begin{subfigure}{0.5\columnwidth}
        \centering
        \includegraphics[width=0.98\textwidth]{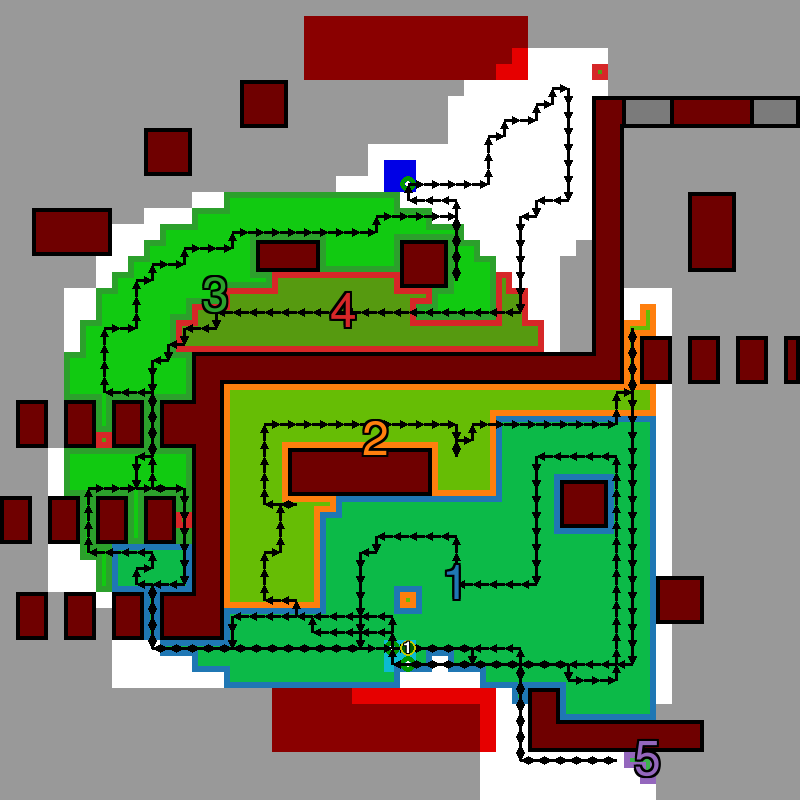}
        \caption{Multi10 agent: 617 steps}\label{fig:multi10_border}
    \end{subfigure}%
    \begin{subfigure}{0.5\columnwidth}
        \centering
        \includegraphics[width=0.98\textwidth]{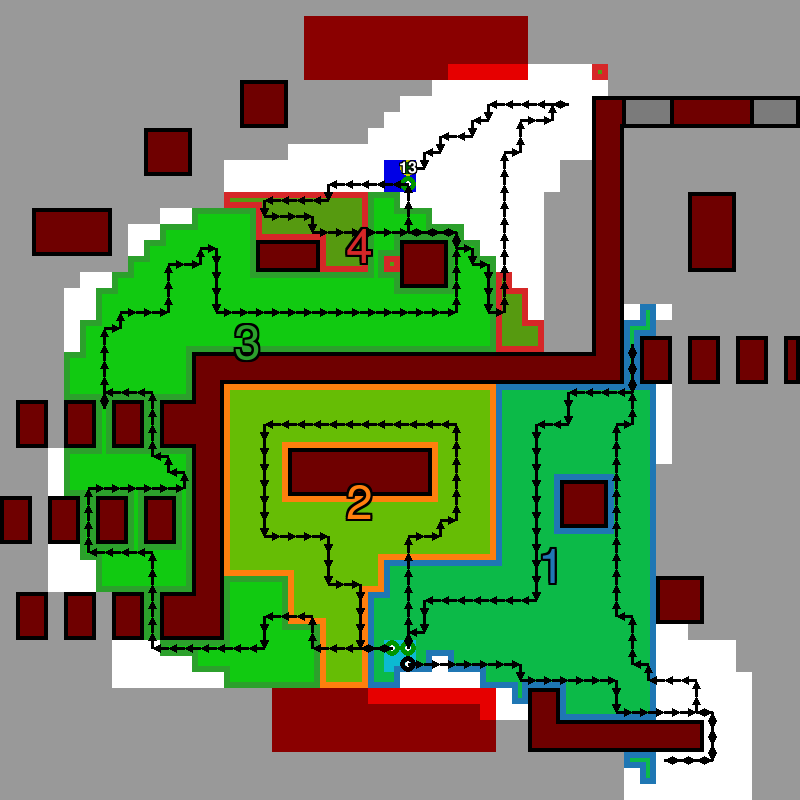}
        \caption{Specialized agent: 435 steps}\label{fig:border_border}
    \end{subfigure}%
    }
    \caption{Trajectories generated by the Multi10, and specialized Border agents on the Border map as an example scenario.}
    \label{fig:border_comp}
\end{figure}

Finally, to evaluate generalization capabilities on a challenging previously unseen map, we tested the agents on the Border map (see Table~\ref{tab:comparison}). Although the map was designed to be more challenging than the maps encountered during training, the Multi10 agent can solve the task in around one third of evaluation scenarios, with the Multi3 agent showing slightly lower performance. As seen from the example trajectory in Figure~\ref{fig:border_comp}, if the Multi10 agent manages to solve the scenario, the resulting decomposition still shows a similar strategy to the highly optimized one of the specialized Border agent. As expected, the specialized agent can also solve the scenario more efficiently than the Multi10 agent and the heuristic, as indicated by a negative RPD value.

\subsubsection{Specialization-Generalization Trade-Off:}
The proposed DRL approach allows the agents to learn efficient paths, outperforming the heuristic and solving multiple maps. As expected, the more specialized an agent is, the better its performance on the in-distribution maps. Similarly, the less specialized the agent is, the better its performance on the out-of-distribution maps. In future work, we will further study the generalization capabilities and explore one-shot and few-shot generalization techniques to improve generalization to unseen maps and to improve the performance of generalizing agents.
\section{Conclusion}
\label{sec:conclusion}

Coverage path planning for UAVs is a problem that is relevant to a large number of application domains. DRL-based solution approaches have been underexplored, especially in the case of power-constrained CPP with recharge. We have introduced a novel PPO-based global-local map DRL approach and provide insights for the DRL algorithm and neural network model design for the challenges of this particular problem. Utilizing a model-based action masking approach, we also guarantee the safety of generated paths. We showed that our approach can generalize path planning over multiple maps with different characteristics and even solve previously unseen maps. Through the detailed evaluation and ablation studies, we hope to provide some guidelines on what works best for developing and implementing DRL-based solutions for long-horizon problems. As not much software is publicly available in this context, we also hope that open-sourcing our software framework with an OpenAI gym-API will prove beneficial for the research community working on the same or related problems.

As expected and shown in this study, we observed a better generalization ability to unseen maps for agents exposed to a larger number of maps during training. In future work, we plan to investigate whether training on even more maps would further improve the generalization ability. We plan on employing a procedural map generator to handle the potentially large requirement of varying maps. At the same time, we observed that the highest performance on a particular map was achieved by a specialized agent only trained on the respective map, which came at the price of dismal generalization ability to unseen maps. To achieve the best combination of specific performance and generalization ability, we also plan to study learning schemes based on the idea of one-shot or few-shot generalization or methods of fine-tuning in the future.
 
\bibliography{IEEEabrv,bib}
\bibliographystyle{IEEEtran}

\balance

\end{document}